\documentclass[11pt]{article}

\PassOptionsToPackage{table}{xcolor}
\usepackage[preprint]{acl}

\usepackage{times}
\usepackage{latexsym}
\usepackage[T1]{fontenc}
\usepackage[utf8]{inputenc}
\usepackage{microtype}
\usepackage{graphicx}
\usepackage{tikz}
\usetikzlibrary{arrows.meta}
\usepackage{pgfplots}
\usepgfplotslibrary{groupplots}
\pgfplotsset{compat=1.18}

\usepackage{url}
\usepackage{booktabs}
\usepackage{multirow}
\usepackage{amsmath}
\usepackage{amsfonts}
\usepackage{amssymb}
\usepackage{amsthm}
\usepackage{nicefrac}
\newtheorem{result}{Result}
\theoremstyle{definition}

\theoremstyle{remark}

\title{Competence-Based Evaluation of Logical Reasoning in Large Language Models}

\author{
  Jize Jiang \qquad Chengxiang Zhai \\
  University of Illinois Urbana-Champaign \\
  \texttt{jizej2@illinois.edu} \qquad \texttt{czhai@illinois.edu}
}
\hypersetup{pdfauthor={Jize Jiang and Chengxiang Zhai}}

\hypersetup{
  pdftitle={Competence-Based Evaluation of Logical Reasoning in Large Language Models}
}

\begin{document}
\maketitle

\begin{abstract}
Recent advances in large language models (LLMs) have brought impressive performance across diverse tasks; however, shortcomings remain in logical reasoning, including inconsistencies under linguistic variation. To improve the reasoning capacity of LLMs, we must be able to accurately measure their reasoning capacity. Current benchmarks primarily report single accuracy numbers on selected reasoning tasks. Such a performance-based evaluation methodology says little about a model's generalization capacity in reasoning. To address this limitation, we argue for a different evaluation paradigm, called \emph{competence-based evaluation} (CBE), which tests whether models reliably apply the same competence across a family of problems. A parameterized generator scales complexity while keeping the required competence fixed, and matched answer-preserving presentations test correctness beyond one formulation. We instantiate CBE on elementary ordering tasks, scaling input size while varying presentations of the same ordering rule. Across six GPT and Qwen models, performance is reasonably good over bounded complexity ranges, but even strong models fail the competence condition across the tested family. 
Chain-of-thought and GRPO extend strong performance without resolving all tested failures; supervised LoRA shows little held-out transfer. 
CBE provides diagnostic profiles and an evaluation protocol that can grow with future models.

\end{abstract}

\section{Introduction}
\label{sec:intro}
Recent large language models (LLMs) achieve strong performance on many reasoning benchmarks, yet growing evidence suggests that high benchmark accuracy does not necessarily imply robust reasoning ability \citep{song2026largelanguagemodelreasoning}. A single accuracy number on a static test set conflates many things: language understanding, memorization, shortcut exploitation, and genuine reasoning. It tells us little about \emph{what a model can actually do}. Recent work has made this conflation increasingly hard to ignore. Benchmarks designed to test logical reasoning can reward other strategies: \citet{chung2025divlogiceval} report higher GPT-3.5 scores on ReClor and LogiQA2 when the prompt asks it to avoid logical reasoning, illustrating that accuracy alone need not diagnose the intended competence. Models that appear to reason well on standard test sets show systematic drops under controlled perturbations \citep{mirzadeh2024gsmsymbolic,li2024gsmplus,sanyal2022robustlr}, and reasoning models can collapse as complexity grows \citep{pmlr-v267-lin25i,shojaee2025illusion}. The natural reading is that reasoning is brittle, but this conflates two questions: whether the model fails to derive a conclusion from a fixed problem representation, or constructs a different representation in the first place. Measuring performance on a fixed test set is insufficient to tell these apart, and therefore insufficient to characterize a model's reasoning capability.

We argue for a different evaluation paradigm: \textbf{competence-based evaluation}. Drawing on Chomsky's distinction between knowledge of a language and its use \citep{chomsky1965aspects}, we adapt the competence/performance distinction to reasoning: underlying task knowledge and observed behavior on specific instances. CBE tests reliable application of that knowledge through an explicit behavioral condition. We also draw on the cognitive-science account given by \citet{stenning2008human} for human reasoning, which distinguishes \emph{reasoning to an interpretation}, establishing a problem's representation and the logic governing it, from \emph{reasoning from an interpretation}, deriving conclusions within that form. This motivates separating the task's latent structure from its realization in language.

A competence claim concerns the family of problems governed by the same rule, including larger instances and equivalent formulations. Exhaustively enumerating that family is infeasible. Concretely, instead of measuring accuracy on a fixed test set, we define a \emph{parameterized problem generator} for a target reasoning competence and test a necessary behavioral condition: the model must answer correctly across the generated instances and their answer-preserving presentations. \textbf{The evaluation target is a problem family whose complexity can grow while the required competence remains the same.} The generator exposes the assumptions of that family and permits evaluation at new scales under the same criterion. A finite experiment can expose violations of the condition; success supports only the evaluated coverage.

CBE characterizes the model's \emph{competence boundary}, the region in the generator's parameter space where correct, invariant behavior deteriorates and exposes biases that aggregate accuracy can conceal. By holding logical form fixed and varying surface form across matched presentations, the evaluation produces a \emph{competence profile} rather than a single score. Paired outputs reveal whether correct answers survive equivalent formulations. They do not by themselves locate a failure in either internal reasoning stage.

This approach yields several advantages at once. Fresh generation reduces dependence on particular published test items, although familiarity with the generator or learned shortcuts remains possible. It produces richer diagnostics than a single accuracy number: the competence profile reveals \emph{where} a model fails and \emph{under which conditions}, including problem size, distractors, and equivalent reformulations. Additionally, as models improve, the evaluation naturally extends by pushing the parameter range further, rather than requiring an entirely new benchmark. Existing generators and consistency tests supply important precedents (\S\ref{sec:positioning}); CBE makes the target competence, its family, and its correctness condition explicit within one evaluation design.

We instantiate the competence-based evaluation framework on \emph{ordering tasks}, a fundamental reasoning primitive involving transitivity and comparison. Ordering is well suited as a first instantiation: the ordering rule is unchanged for any finite number of entities, so an extremely simple task supports a scalable problem family. The generator controls the latent order and renders matched presentations that vary premise order, converse phrasing, names order, and relation vocabulary (Figure~\ref{fig:generation}). Failure on this elementary family violates the specified competence condition. Chain length scales input and presentation load; endpoint queries admit a shortcut and do not isolate multistep inference (\S\ref{sec:gen_ordering}). Appendix~\ref{app:horn} illustrates adoption of the framework by Horn-clause inference.

Our results show why the distinction between performance and competence matters. Even strong models fail the CBE condition across the tested ordering family, while performing reasonably well over bounded complexity ranges. On front/behind at $n=1024$, GPT-4.1 retains 77.5\% original-prompt accuracy but answers both matched presentations correctly on only 56.8\% of valid pairs. 
Vocabulary changes alter model rankings. Chain-of-thought extends strong performance, and GRPO improves held-out transfer, but the tested interventions leave substantial failures. Improved scores can therefore reflect an expanded range of successful performance without demonstrating competence across the family.

\paragraph{Contributions.}
(1) \textbf{Competence-based evaluation as a general framework.} We formalize the shift from performance on selected instances to a competence condition on a specified problem family, through a parameterized task generator, presentation set, model, correctness criterion, and competence condition (\S\ref{sec:framework}).
(2) \textbf{The CBE Ordering Benchmark.} We instantiate the framework on an elementary ordering task with precisely controlled latent structure and matched presentations. Its size parameter extends the evaluation while retaining the same ordering rule (\S\ref{sec:instantiation}).
(3) \textbf{A multidimensional competence analysis.} We evaluate six base models and reasoning/training interventions, showing where correctness fails under complexity and presentation changes (\S\ref{sec:results}--\S\ref{sec:training}). 
The analysis distinguishes gains in aggregate performance from gains in the range of correct, invariant behavior, and provides diagnostic evidence that can guide future model development.

\section{Related Work}
\label{sec:positioning}
We position CBE relative to fixed test sets, generated complexity evaluations, and invariance tests. The distinction is the evaluation target: a defined competence over a scalable problem family, with correctness required across equivalent presentations. Appendix~\ref{app:positioning} compares the designs by axis.

\subsection{Natural-Language Reasoning Benchmarks}

ReClor, LogiQA, AR-LSAT, and FOLIO assess reasoning on curated or annotated problems \citep{yu2020reclor,liu2020logiqa,zhong-etal-2022-analytical,han-etal-2024-folio}. DivLogicEval uses counterintuitive content to reduce shortcuts \citep{chung2025divlogiceval}, while BIG-bench \texttt{logical\_deduction} generates linear-ordering problems at several sizes \citep{srivastava2023bigbench}. These datasets provide valuable samples of task performance. CBE organizes correctness across equivalent presentations and sizes around an explicit competence condition on a specified problem family.

\subsection{Generated and Complexity-Scaled Evaluation}

RuleTaker, ProofWriter, CLUTRR, and PrOntoQA generate deductive or relational instances with controlled structure or depth \citep{clark2020transformers,tafjord2021proofwriter,sinha2019clutrr,saparov2023greedy}. DyVal, functional benchmarks, NPHardEval, and ZebraLogic likewise generate instances or vary complexity \citep{zhu2024dyval,srivastava2024functional,fan2024nphardeval,pmlr-v267-lin25i}. ReCogLab is especially close: it generates relational reasoning, including transitive comparison, with entity count, reasoning length, and textual features as parameters \citep{liu2025recoglab}. These works establish generation and complexity scaling as useful evaluation tools. CBE couples that control to an explicit competence condition on matched presentations and evaluates the resulting profile across the family.

\subsection{Invariance and Diagnostic Evaluation}

CheckList and metamorphic testing prescribe relations between outputs on transformed inputs \citep{ribeiro2020checklist,chen1998metamorphic,segura2016survey,zhou2026lgmt}. ParaRel reports consistency and \emph{Consistent-Acc}, correctness across all paraphrases \citep{elazar2021pararel}; BECEL shows that equal accuracy can hide pairwise inconsistency \citep{jang2022becel}. RobustLR, GSM-Symbolic, GSM-Plus, and premise-order studies expose sensitivity to controlled changes \citep{sanyal2022robustlr,mirzadeh2024gsmsymbolic,li2024gsmplus,chen2024premise}. These are direct precedents for the paired measurements. CBE's distinction is to make such requirements part of the competence definition and trace their satisfaction as complexity grows within one fixed task family. A robustness score on a chosen set of examples, or an accuracy curve alone, leaves that combined question unanswered.

\section{Competence-Based Evaluation}
\label{sec:framework}
\subsection{A Competence and Its Problem Family}
\label{sec:estimand}

CBE has five components: a parameterized generator $G_\theta$, a presentation set $\Pi$, a model $\mathcal M$, a correctness criterion, and a competence condition. The target competence fixes the problem rule. Parameters $\theta$ vary instance size or presentation load without changing that rule. The generator draws a latent instance $z$; an exact oracle supplies $y^*(z)$. Each $\pi_k\in\Pi(z)$ preserves the answer, $y^*(\pi_k(z))=y^*(z)$.

Write $C_k=\mathbf1[\mathcal M(\pi_k(z))=y^*(z)]$. The necessary behavioral condition for competence on an instance is
\begin{equation}
C_0=C_1=\cdots=C_K=1.
\end{equation}
The family-level condition requires this behavior throughout the specified instance and complexity space. A finite evaluation can expose failures of this operational condition; it cannot certify competence exhaustively or identify their internal cause. Equivalent presentations prevent success on one formulation from serving as the entire competence test.

\subsection{Profiles of Correctness and Invariance}
\label{sec:identifiability}
\label{sec:profiles}

For two presentations with valid binary answers, let $R$ be the probability of both correct, $D$ both wrong, and $U$ exactly one correct. These robust-correct, shared-wrong, and presentation-dependent masses satisfy
\begin{equation}
R+D+U=1,\quad \bar A=R+U/2,\quad \iota=R+D,
\end{equation}
where $\bar A$ is mean accuracy and $\iota$ is invariance. The \emph{competence profile} is $\theta\mapsto(R,D,U)(\theta)$. Accuracy can conceal offsetting changes, and agreement can include shared errors. Appendix~\ref{app:proofs} gives the elementary identities and marginal bounds.

We also report $\kappa_n=(\iota_n-\iota_e)/(1-\iota_e)$, where $\iota_e=q^2+(1-q)^2$ and $q$ is the pooled yes-rate within $n$. This discounts a common-bias independent-answer reference. It is undefined for constant outputs and does not by itself identify independence or a failure mechanism. The summary $\bar\kappa$ averages $\kappa_n$ over the grid.

\subsection{Readout and Inference}
\label{sec:testing}

Profiles locate regions of deterioration, rather than a single threshold crossing. Accuracy counts invalid responses wrong on all prompts; pair metrics use jointly parsed answers. All profile identities use common support. McNemar tests marginal accuracy differences, including format failures, with Holm correction. Paired bootstrap intervals resample matched instances. Decoding settings and statistical details appear in Appendix~\ref{app:details}.

\section{An Instantiation of CBE: Ordering}
\label{sec:instantiation}
\subsection{The Ordering Family}
\label{sec:gen_ordering}

An instance is a total order over $n$ named entities, described by its $n-1$ adjacent comparisons. A yes/no query concerns the two endpoints. Query direction and gold answer form four balanced cells. Names and premises are shuffled independently of the latent order. The ordering rule is unchanged as $n$ increases. We evaluate $n\in\{4,8,\ldots,1024\}$ in the base study. 
Figure~\ref{fig:generation} shows generation and paired scoring.

\begin{figure*}[t]
\centering
\resizebox{\textwidth}{!}{
\begingroup
\definecolor{original}{HTML}{0072B2}
\definecolor{equivalent}{HTML}{009E73}
\definecolor{ink}{HTML}{263238}
\begin{tikzpicture}[x=1pt,y=1pt,font=\fontsize{9}{10.5}\selectfont,
  every node/.style={text=ink},arr/.style={-{Stealth[length=4pt]},draw=ink,line width=.65pt},
  box/.style={rounded corners=3pt,line width=.6pt,draw=ink!50,fill=ink!3}]
\node[anchor=west,font=\bfseries\fontsize{10}{11}\selectfont] at (0,146) {1. Sample latent order};
\node[anchor=west,font=\bfseries\fontsize{10}{11}\selectfont] at (118,146) {2. Render matched prompts};
\node[anchor=west,font=\bfseries\fontsize{10}{11}\selectfont] at (335,146) {3. Score paired answers};
\draw[box] (0,8) rectangle (94,136);
\node[anchor=north,align=center] at (47,129) {Illustration: $n=4$};
\node (a) at (47,104) {Amy};
\node (b) at (47,80) {Bob};
\node (c) at (47,56) {Cam};
\node (d) at (47,32) {Dan};
\draw[arr] (a) -- (b);
\draw[arr] (b) -- (c);
\draw[arr] (c) -- (d);
\node[anchor=south,font=\fontsize{8}{9}\selectfont] at (47,10) {front $\longrightarrow$ back};
\draw[arr] (97,101) -- (115,101);
\draw[arr] (97,42) -- (115,42);
\draw[box,draw=original,fill=original!4] (118,77) rectangle (310,136);
\node[anchor=north west,text=original,font=\bfseries] at (123,133) {Original $\pi_0$};
\node[anchor=north west,align=left,font=\fontsize{8.5}{10}\selectfont] at (123,120)
  {Names: Cam, Amy, Dan, Bob\\
   Cam is in front of Dan.\\
   Amy is in front of Bob.\\
   Bob is in front of Cam.};
\draw[box,draw=equivalent,fill=equivalent!4] (118,8) rectangle (310,67);
\node[anchor=north west,text=equivalent,font=\bfseries] at (123,64) {Equivalent $\pi_1$};
\node[anchor=north west,align=left,font=\fontsize{8.5}{10}\selectfont] at (123,51)
  {Names: Bob, Dan, Amy, Cam\\
   Bob is behind Amy.\\
   Bob is in front of Cam.\\
   Dan is behind Cam.};
\draw[arr] (313,106) -- (333,106);
\draw[arr] (313,38) -- (333,38);
\draw[box] (335,8) rectangle (455,136);
\node[anchor=north,align=center,font=\bfseries] at (395,129) {Model outputs\\$(\hat y_0,\hat y_1)$};
\node[anchor=west,align=left] at (341,88) {$R$: yes / yes\\Both correct};
\node[anchor=west,align=left] at (341,59) {$D$: no / no\\Both wrong};
\node[anchor=west,align=left] at (341,30) {$U$: yes / no or no / yes\\One correct};
\node[anchor=north west,align=left] at (0,-1)
  {\textbf{Shared query:} Is Amy in front of Dan? \quad \textbf{Gold:} yes};
\node[anchor=north west,align=left,font=\fontsize{8.5}{10}\selectfont] at (0,-17)
  {Swap arguments + use the converse phrase; shuffle premises and names independently.};
\node[anchor=north west,align=left,font=\bfseries\fontsize{9}{10.5}\selectfont] at (0,-33)
  {Repeat over balanced query cells and chain lengths $n$ $\longrightarrow$ competence profile $(R,D,U)_n$.};
\end{tikzpicture}
\endgroup}
\caption{Ordering generation and paired evaluation (illustrative instance). A latent order supplies adjacent premises and a gold endpoint answer. Equivalent prompts swap arguments with converse phrasing and reshuffle names and premises, preserving the context and query. Paired answers determine both-correct ($R$), both-wrong ($D$), or exactly-one-correct ($U$) outcomes; repetition across instances and chain lengths yields the competence profile.}
\label{fig:generation}
\end{figure*}

The original presentation uses the forward phrase, such as “Amy is in front of Bob”. Its equivalent rewrites selected links as “Bob is behind Amy” and reshuffles the names list and premises. Controls separate reshuffling from rewriting. A catalog renders the same latent instances in 31 vocabularies across six relation families. 
Negation preserves answers under the stated strict total order; distractors over disjoint entities add irrelevant input.

Ordering supplies a scalable test of reliable application of the same rule. Here $n$ measures input size and presentation load: endpoint queries admit a source/sink shortcut using an incident premise. Success need not reconstruct the chain. Failures violate the specified competence condition, but do not isolate a multistep inference limit or distinguish inference failures from retrieval and presentation handling. Appendix~\ref{app:horn} illustrates adoption of CBE for Horn-clause inference.

\subsection{Evaluation Setup and Models}
\label{sec:protocol}

The base evaluation uses seed-1 instances, 200 pairs per $n$, 50 per query cell, randomly selected rewrite positions, and identical latent instances across models, vocabularies, and presentation controls. 
The six base models are GPT-4.1, GPT-4.1-mini, GPT-5, GPT-5-mini, Qwen3.5-27B, and Qwen3.5-9B; GPT-5 uses minimal reasoning effort and Qwen thinking is disabled. 
Qwen models cover the full catalog and GPT models six spatial-line vocabularies. 
Prompts request yes/no answers under provider-default sampling. Caps and parsing rules appear in Appendix~\ref{app:details}, and decoding-variability controls in Appendix~\ref{app:repeat_agreement}.

Reasoning-effort, distractor, and negation studies use 200 matched pairs per complexity level. 
All effort levels receive identical prompts within each condition. The separate $n=2048$ block changes the presentation distribution, so its failures cannot be attributed solely to length. Additional models and earlier evaluations are in Appendices~\ref{app:per_n} and \ref{app:legacy}.

\section{Competence on Ordering}
\label{sec:results}
We read performance through accuracy, invariance, and robust correctness. Matched comparisons vary input size, vocabulary, presentation factors, and reasoning effort.

\subsection{Bounded Performance on Front/Behind}
\label{sec:base_pair}

\begin{table*}[t]
\centering
\small
\setlength{\tabcolsep}{5pt}
\begin{tabular}{lrrrrr@{\hspace{15pt}}rrrr}
\toprule
 & \multicolumn{5}{c}{front/behind} & \multicolumn{4}{c}{left/right} \\
\cmidrule(lr){2-6}\cmidrule(lr){7-10}
Model & Acc.Org & Acc.Eqv & Inv. & $R$ & $\bar\kappa$ & Acc.Org & Acc.Eqv & Inv. & $\bar\kappa$ \\
\midrule
GPT-4.1      & 93.1 & 89.8 & 87.9 & 85.4 & 0.75 & 70.9 & 83.8 & 77.7 & 0.33 \\
GPT-4.1-mini & 75.9 & 71.4 & 72.5 & 59.9 & 0.36 & 79.1 & 76.6 & 71.6 & 0.42 \\
GPT-5        & 94.1 & 87.1 & 85.7 & 83.6 & 0.69 & 80.4 & 90.0 & 73.5 & 0.45 \\
GPT-5-mini   & 70.0 & 72.0 & 67.1 & 54.6 & 0.26 & 70.5 & 71.4 & 66.0 & 0.26 \\
Qwen3.5-27B  & 81.4 & 72.9 & 72.1 & 63.2 & 0.37 & 86.2 & 83.3 & 77.4 & 0.54 \\
Qwen3.5-9B   & 55.6 & 53.3 & 57.3 & 33.2 & 0.00 & 64.2 & 66.1 & 66.9 & 0.20 \\
\bottomrule
\end{tabular}
\caption{Base ordering results pooled over $n\in\{4,8,\ldots,1024\}$ (200 pairs per level). Acc.Org/Eqv score all prompts, counting invalid answers wrong; invariance (Inv.) and robust correctness ($R$) condition on jointly parsed answers. $\bar\kappa$ averages chance-corrected invariance over the complexity grid. Percentages except $\bar\kappa$. Left/right renders the same latent instances with a different vocabulary.}
\label{tab:base_pair}
\end{table*}

\begin{figure*}[t]
\centering
\includegraphics[width=\textwidth]{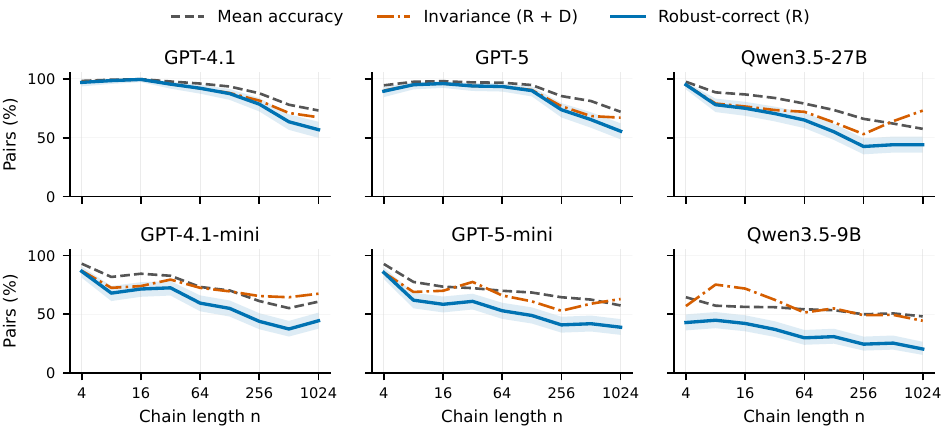}
\caption{Ordering competence profiles for the six base models on front/behind (200 pairs per chain length). Mean accuracy and both paired measures use the same valid-pair support. Robust correctness $R$ requires both presentations to be correct; invariance $R+D$ also includes both-wrong pairs. Shading gives pointwise 95\% Wilson intervals for $R$. Accuracy can decline less than robust correctness, while shared errors can keep invariance high. Chance-corrected profiles for both vocabularies appear in Appendix~\ref{app:per_n}.}
\label{fig:competence_profiles}
\end{figure*}

GPT-4.1 and GPT-5 perform well on shorter front/behind chains. Through $n=128$, original accuracy remains at least 94.0\% and invariance at least 87.9\%. At $n=1024$, original accuracy is 77.5\% and 76.0\%, but invariance is only 67.3\% and 67.0\%, with $\kappa_n=0.30$ and $0.16$. 
The smaller GPT models and Qwen3.5-27B show earlier deterioration: invariance drops from 87.0 to 72.5, 85.5 to 69.0, and 95.0 to 79.0 between $n=4$ and $8$, respectively. 
These profiles support reasonably good performance over bounded ranges, with clear failures of the competence condition across the full tested range.

\paragraph{Accuracy understates paired deterioration.}
Figure~\ref{fig:competence_profiles} makes the difference explicit. On the common valid-pair support, GPT-4.1's mean accuracy falls from 98.3\% at $n=4$ to 73.1\% at $n=1024$, while invariance falls from 97.5\% to 67.3\% and robust correctness from 97.0\% to 56.8\%. 
At the latter level, nearly a third of pairs are presentation-dependent ($U=32.7\%$). 
The pattern is not a universal ordering of decline rates: weak models can acquire shared wrong answers that increase agreement. Qwen3.5-9B, for example, reaches 75.3\% raw invariance at $n=8$ but has $\bar\kappa=0.00$ over the grid. 
Correctness and invariance must therefore be interpreted jointly.

Marginal gaps also hide symmetric instability. GPT-5-mini has pooled original/equivalent accuracy of 70.0/72.0\% and no significant McNemar difference, yet $R/D/U=54.6/12.6/32.9\%$. 
Its discordant counts are $278/314$: substantial correct-to-wrong changes are offset by changes in the reverse direction. 
A small accuracy gap therefore does not establish stable task performance.

\subsection{Vocabulary Changes the Profile}
\label{sec:relation_swap}

Switching only front/behind to left/right changes the rankings (Table~\ref{tab:base_pair}). GPT-4.1's original accuracy falls from 93.1\% to 70.9\% and $\bar\kappa$ from 0.75 to 0.33. Qwen3.5-27B instead gains, from $\bar\kappa=0.37$ to $0.54$. 
For GPT-4.1 on left/right, the original presentation elicits “yes” on 78\% of prompts despite balanced gold answers; equivalent-presentation accuracy exceeds original accuracy at every $n\ge8$. 
This is behavioral evidence of a vocabulary-dependent answer bias.

The full catalog shows that the base pair alone is insufficient to characterize the family. Qwen3.5-27B's family-mean original accuracy and $\bar\kappa$ are 79.6\%/0.38 for spatial-line relations and 87.2\%/0.58 for comparatives; the corresponding 9B values are 61.0\%/0.09 and 75.7\%/0.30. 
Across relations, original accuracy spans 27 points and invariance 23 points for a single model. 
Nonce comparative terms can be handled well (27B accuracy/invariance 87.1/87.6\%), while nonce direction terms yield only 54.6\% equivalent accuracy and $\bar\kappa\le0.03$ for every tested model, even with the converse rule supplied. 
Stating the rule is insufficient in this condition; these results do not establish that lexical familiarity is necessary for every way of learning the task.

\subsection{Separating Presentation Factors}
\label{sec:forms}

Reshuffling alone changes GPT accuracy by at most 1.4 points. Fully rewriting premises costs GPT-4.1 and GPT-5 a further 9.1/11.2 accuracy points and 12.4/9.7 invariance points relative to reshuffling alone (Table~\ref{tab:forms}). 
The effects of reordering and converse phrasing differ, although both preserve the answer.

Additional matched controls add distractors or use negation (Table~\ref{tab:dose_negation}). For GPT-5-mini, distractors produce 14.8\% correct-to-wrong changes and 14.6\% changes in the reverse direction, leaving aggregate performance almost unchanged. 
Negation reduces agreement in all four tested models; for Qwen3.5-27B, syntactic and negated-presentation agreement are 75.2 and 60.2\%. 
Both controls reinforce the need to follow matched instances.

\subsection{Reasoning Extends the Strong-Performance Range}
\label{sec:reasoning}

\begin{table}[t]
\centering
\scriptsize
\setlength{\tabcolsep}{3pt}
\begin{tabular}{lrrrrr}
\toprule
Effort & Acc.Org & Acc.Eqv & Inv. & $\bar\kappa$ & tokens org/eqv \\
\midrule
\multicolumn{6}{l}{\emph{pooled over $n\in\{4,\dots,1024\}$}} \\
\texttt{none}   & 77.1 & 82.6 & 79.2 & 0.47 & 0/0 \\
\texttt{low}    & 97.7 & 96.9 & 96.0 & 0.91 & 1,014/1,545 \\
\texttt{medium} & 99.4 & 98.8 & 98.4 & 0.97 & 1,641/2,718 \\
\texttt{high}   & 99.9 & 99.4 & 99.4 & 0.99 & 2,052/3,455 \\
\midrule
\multicolumn{6}{l}{\emph{$n=2048$ block (200 pairs; $\kappa_{2048}$ in the $\bar\kappa$ column)}} \\
\texttt{none}   & 67.0 & 67.5 & 60.5 & 0.08 & 0/0 \\
\texttt{low}    & 64.5 & 53.0 & 82.5 & 0.05 & 3,623/910 \\
\texttt{medium} & 85.0 & 61.5 & 68.5 & 0.13 & 10,230/7,412 \\
\texttt{high}   & 94.0 & 65.0 & 67.0 & 0.21 & 15,130/16,641 \\
\bottomrule
\end{tabular}
\caption{GPT-5.4 by reasoning effort on front/behind (200 pairs per $n$). Accuracy (\%) on both presentations, paired invariance, chance correction, and mean reasoning tokens (original/equivalent). Every effort level uses the same prompts within each condition. The $n=2048$ block uses a distinct presentation distribution (Appendix~\ref{app:details}). No shown answer was truncated.}
\label{tab:reasoning}
\end{table}

\begin{figure*}[t]
\centering
\resizebox{\textwidth}{!}{\begingroup
\definecolor{curve0}{HTML}{666666}
\definecolor{curve1}{HTML}{0072B2}
\definecolor{curve2}{HTML}{D55E00}
\definecolor{curve3}{HTML}{009E73}
\begin{tikzpicture}
\begin{groupplot}[group style={group size=2 by 1,horizontal sep=36pt},
width=212pt,height=124pt,xmode=log,log basis x=2,
tick label style={font=\fontsize{8}{9}\selectfont},
label style={font=\fontsize{9}{10}\selectfont},
title style={font=\fontsize{9}{10}\selectfont},
xlabel={Chain length $n$},xmin=3.8,
xmax=2600,xtick={4,16,64,256,1024,2048},xticklabels={4,16,64,256,1k,2k},
axis lines=left,ymajorgrids=true,grid style={black!12},
legend style={font=\fontsize{8}{9}\selectfont,draw=none,at={(1.16,1.28)},
anchor=south,legend columns=4,column sep=6pt},
every axis plot/.append style={line width=.9pt,mark size=1.4pt}]
\nextgroupplot[ymin=40,ymax=103,ytick={50,75,100},title={Mean accuracy},ylabel={Accuracy (\%)}]
\addplot[color=curve0,dotted,mark=o] coordinates {(4,97.75000000) (8,90.75000000) (16,84.00000000) (32,81.75000000) (64,77.25000000) (128,74.00000000) (256,69.75000000) (512,71.25000000) (1024,72.00000000)};
\addlegendentry{none}
\addplot[color=curve0,dotted,mark=o,only marks,forget plot] coordinates {(2048,67.25000000)};
\addplot[color=curve1,dashed,mark=square*] coordinates {(4,100.00000000) (8,100.00000000) (16,100.00000000) (32,99.75000000) (64,99.50000000) (128,99.00000000) (256,99.00000000) (512,95.75000000) (1024,83.00000000)};
\addlegendentry{low}
\addplot[color=curve1,dashed,mark=square*,only marks,forget plot] coordinates {(2048,58.75000000)};
\addplot[color=curve2,dashdotted,mark=diamond*] coordinates {(4,100.00000000) (8,100.00000000) (16,100.00000000) (32,99.75000000) (64,98.50000000) (128,99.00000000) (256,100.00000000) (512,99.00000000) (1024,95.75000000)};
\addlegendentry{medium}
\addplot[color=curve2,dashdotted,mark=diamond*,only marks,forget plot] coordinates {(2048,73.25000000)};
\addplot[color=curve3,solid,mark=triangle*] coordinates {(4,100.00000000) (8,100.00000000) (16,100.00000000) (32,99.50000000) (64,100.00000000) (128,100.00000000) (256,99.50000000) (512,99.25000000) (1024,98.75000000)};
\addlegendentry{high}
\addplot[color=curve3,solid,mark=triangle*,only marks,forget plot] coordinates {(2048,79.50000000)};
\draw[black!45,dashed] (axis cs:1448,40) -- (axis cs:1448,103);
\nextgroupplot[ymin=-.3,ymax=1.03,ytick={-.2,0,.5,1},title={Chance-corrected invariance},ylabel={$\kappa_n$}]
\draw[black!40,thin] (axis cs:4,0) -- (axis cs:2600,0);
\addplot[color=curve0,dotted,mark=o] coordinates {(4,0.90981738) (8,0.69972302) (16,0.46524064) (32,0.57313028) (64,0.43251679) (128,0.45175439) (256,0.32173982) (512,0.24606138) (1024,0.11212582)};
\addplot[color=curve0,dotted,mark=o,only marks,forget plot] coordinates {(2048,0.08072727)};
\addplot[color=curve1,dashed,mark=square*] coordinates {(4,1.00000000) (8,1.00000000) (16,1.00000000) (32,0.98999975) (64,0.97999800) (128,0.95998399) (256,0.95998399) (512,0.82979149) (1024,0.50248756)};
\addplot[color=curve1,dashed,mark=square*,only marks,forget plot] coordinates {(2048,0.04884843)};
\addplot[color=curve2,dashdotted,mark=diamond*] coordinates {(4,1.00000000) (8,1.00000000) (16,1.00000000) (32,0.98999975) (64,0.94000000) (128,0.95998399) (256,1.00000000) (512,0.95998399) (1024,0.86973622)};
\addplot[color=curve2,dashdotted,mark=diamond*,only marks,forget plot] coordinates {(2048,0.13028473)};
\addplot[color=curve3,solid,mark=triangle*] coordinates {(4,1.00000000) (8,1.00000000) (16,1.00000000) (32,0.97999800) (64,1.00000000) (128,1.00000000) (256,0.97999800) (512,0.96999325) (1024,0.96999925)};
\addplot[color=curve3,solid,mark=triangle*,only marks,forget plot] coordinates {(2048,0.20663541)};
\draw[black!45,dashed] (axis cs:1448,-.3) -- (axis cs:1448,1.03);
\end{groupplot}
\end{tikzpicture}
\endgroup}
\caption{GPT-5.4 reasoning-effort curves on front/behind (200 matched pairs per level). Mean accuracy and $\kappa_n$ share valid-pair support. Curves end at $n=1024$; isolated $n=2048$ markers, beyond the dashed divider, represent a separate presentation condition (Appendix~\ref{app:details}). More reasoning extends strong performance, while the separate longer-chain condition retains failures.}
\label{fig:reasoning_curves}
\end{figure*}
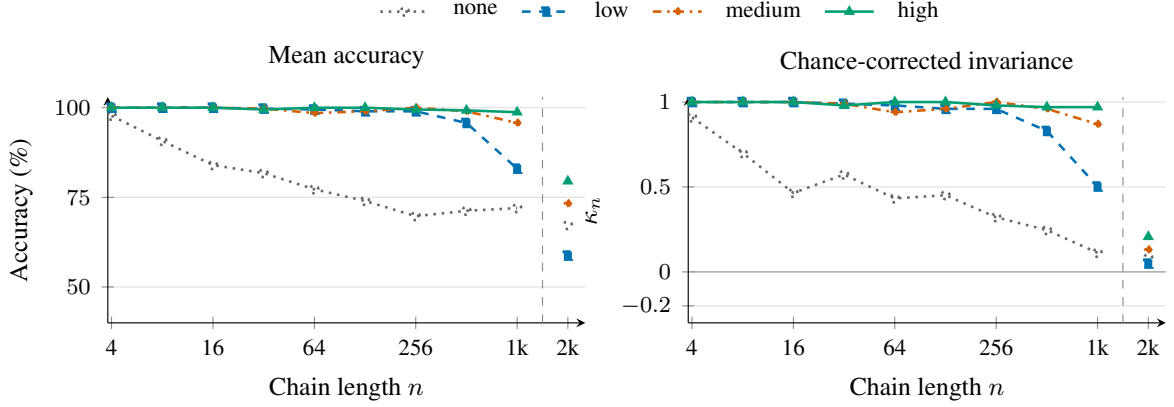

Figure~\ref{fig:reasoning_curves} shows how elicited chain-of-thought changes the complexity profile. On matched front/behind prompts, moving GPT-5.4 from \texttt{none} to \texttt{low} increases original accuracy by 20.6 points and invariance by 16.8 points (95\% paired interval 14.7--18.8). 
Medium and high effort maintain near-perfect aggregate performance through $n=1024$, while low-effort invariance falls to 78.0\% at that level. 
Thus added reasoning extends the range supported by the evaluation.

It does not establish competence across all tested conditions. In the separate $n=2048$ block, high effort attains 94.0\% original accuracy but 65.0\% equivalent accuracy and 67.0\% invariance, without truncation. 
This condition also changes the presentation distribution (Appendix~\ref{app:details}), so the failure cannot be attributed solely to input length. Effort levels remain matched within the condition. Equivalent prompts use fewer reasoning tokens than originals at low and medium effort but more at high effort. 
Neither additional effort nor available completion budget ensures correct, invariant answers under the tested settings.

Qwen3.5-27B thinking results show a different limitation. With a 24k-token budget, accuracy among finished answers is 100.0/99.6\% and invariance among jointly finished pairs is 99.9\%. All-prompt accuracy is 87.0/79.1\%, and strict invariance is 75.4\% when unfinished pairs count as disagreements. 
Conditional success and budget-limited coverage must both be reported.

\section{What Training Changes}
\label{sec:training}
We compare supervised fine-tuning (SFT) and Group Relative Policy Optimization (GRPO) as LoRA interventions on Qwen3.5-9B \citep{hu2022lora,shao2024deepseekmath}. Front/behind is held out. SFT uses auxiliary relation data, paired presentations, or paired presentations with a consistency loss. GRPO training excludes spatial-line vocabularies to evaluate transfer to held-out relation words. 

GRPO samples $G=4$ answers per prompt, with four presentations of one instance in each batch. A sampled answer receives reward
\begin{equation}
r_i=c_i+\tfrac12 b_i,
\end{equation}
where $c_i$ is one for a correct answer and zero otherwise, and $b_i$ is a nonnegative bonus for cross-presentation agreement above a reference derived from the instance's sampled yes/no balance. Unparseable answers receive zero reward. Rewards are centered and scaled within each prompt's sample group to form policy advantages. Appendix~\ref{app:training} gives the full reward definition and training hyperparameters. 

\begin{table}[t]
\centering
\footnotesize\setlength{\tabcolsep}{2pt}
\begin{tabular}{lrrrr}
\toprule
Arm & Acc.Org & Acc.Eqv & Inv. & $\bar\kappa$ \\
\midrule
\multicolumn{5}{l}{\emph{Held-out front/behind}} \\
9B base & 55.6 & 53.3 & 57.3 & 0.00 \\
SFT, cross-relation & 56.4 & 55.0 & 56.6 & $-$0.01 \\
SFT, pairs & 54.5 & 56.6 & 58.0 & 0.00 \\
SFT, pairs + JS & 56.2 & 54.5 & 58.3 & 0.00 \\
GRPO & 81.2 & 75.1 & 73.3 & 0.33 \\
27B base reference & 81.4 & 72.9 & 72.1 & 0.37 \\
\midrule
\multicolumn{5}{l}{\emph{Means over twelve clean held-out relations}} \\
9B base & 61.4 & 60.2 & 60.0 & 0.10 \\
GRPO & 87.6 & 83.7 & 83.5 & 0.60 \\
\bottomrule
\end{tabular}
\caption{Training comparisons ($n=4$ to $1024$, 200 pairs per level). All adapters use LoRA on Qwen3.5-9B. GRPO reports seed 42, selected for higher mean invariance on the twelve clean held-out relations; the other seed is in Appendix~\ref{app:grpo}. JS adds a consistency loss. Accuracy uses all prompts; pair measures use jointly parsed answers. Percentages except $\bar\kappa$.}
\label{tab:training}
\end{table}

\begin{figure*}[t]
\centering
\resizebox{\textwidth}{!}{\begingroup
\definecolor{curve0}{HTML}{666666}
\definecolor{curve1}{HTML}{0072B2}
\definecolor{curve2}{HTML}{D55E00}
\definecolor{curve3}{HTML}{009E73}
\begin{tikzpicture}
\begin{groupplot}[group style={group size=2 by 1,horizontal sep=36pt},
width=212pt,height=124pt,xmode=log,log basis x=2,
tick label style={font=\fontsize{8}{9}\selectfont},
label style={font=\fontsize{9}{10}\selectfont},
title style={font=\fontsize{9}{10}\selectfont},
xlabel={Chain length $n$},xmin=3.8,
xmax=1150,xtick={4,16,64,256,1024},xticklabels={4,16,64,256,1024},
axis lines=left,ymajorgrids=true,grid style={black!12},
legend style={font=\fontsize{8}{9}\selectfont,draw=none,at={(1.16,1.28)},
anchor=south,legend columns=3,column sep=6pt},
every axis plot/.append style={line width=.9pt,mark size=1.4pt}]
\nextgroupplot[ymin=40,ymax=103,ytick={50,75,100},title={Mean accuracy},ylabel={Accuracy (\%)}]
\addplot[color=curve0,dotted,mark=o] coordinates {(4,64.50000000) (8,57.32323232) (16,56.28140704) (32,56.03015075) (64,54.25000000) (128,53.50000000) (256,50.00000000) (512,50.75000000) (1024,48.25000000)};
\addlegendentry{Base 9B}
\addplot[color=curve1,dashed,mark=square*] coordinates {(4,66.83417085) (8,54.77386935) (16,53.51758794) (32,57.78894472) (64,55.61224490) (128,50.76142132) (256,55.77889447) (512,52.76381910) (1024,52.01005025)};
\addlegendentry{SFT + JS}
\addplot[color=curve2,dashdotted,mark=diamond*] coordinates {(4,92.50000000) (8,95.25000000) (16,91.00000000) (32,88.50000000) (64,84.00000000) (128,76.50000000) (256,63.50000000) (512,58.75000000) (1024,53.25000000)};
\addlegendentry{GRPO}
\nextgroupplot[ymin=-.3,ymax=1.03,ytick={-.2,0,.5,1},title={Chance-corrected invariance},ylabel={$\kappa_n$}]
\draw[black!40,thin] (axis cs:4,0) -- (axis cs:1150,0);
\addplot[color=curve0,dotted,mark=o] coordinates {(4,0.00359170) (8,0.17706434) (16,-0.01715955) (32,-0.05235325) (64,-0.05575358) (128,0.06629318) (256,-0.01517792) (512,-0.01002525) (1024,-0.13383896)};
\addplot[color=curve1,dashed,mark=square*] coordinates {(4,0.16130268) (8,0.05068575) (16,0.00066076) (32,-0.05626327) (64,0.11127451) (128,-0.00685080) (256,-0.02577320) (512,-0.10555556) (1024,-0.08868267)};
\addplot[color=curve2,dashdotted,mark=diamond*] coordinates {(4,0.73871973) (8,0.82845178) (16,0.67793881) (32,0.59803035) (64,0.41626409) (128,0.18933333) (256,-0.12390351) (512,-0.11275696) (1024,-0.19794872)};
\end{groupplot}
\end{tikzpicture}
\endgroup}
\caption{Training changes the ordering complexity profile (held-out front/behind, 200 pairs per level). Mean accuracy and $\kappa_n$ share valid-pair support. The selected GRPO run is compared with the base 9B and SFT with a JS consistency loss. Gains at smaller sizes coexist with weak long-chain invariance.}
\label{fig:grpo_curves}
\end{figure*}
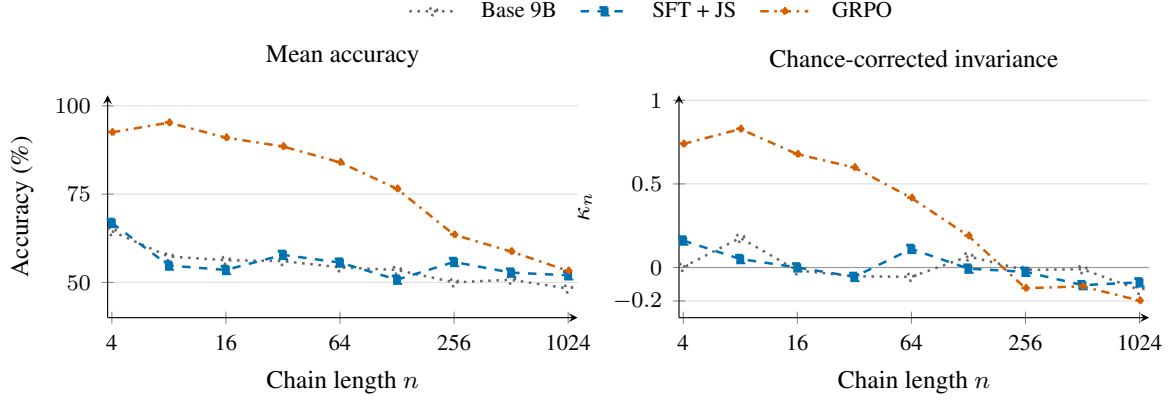

\paragraph{SFT learns the training relations but transfers little.}
The supervised variants reach 100\% on their held-in relations. On front/behind, invariance remains 56.6--58.3\%, compared with 57.3\% for the base model; the cross-relation paired change is $-0.4$ points with interval $[-3.5,+2.5]$. 
The consistency term produces little additional held-out improvement. The larger-model SFT replicate is in Appendix~\ref{app:legacy}.

\paragraph{GRPO improves transfer under the tested recipe.}
Across twelve clean held-out relations, GRPO raises invariance from 60.0\% to 83.5\% and $\bar\kappa$ from 0.10 to 0.60. 
On the nine relations with no partial phrase overlap, invariance increases from 59.9\% to 82.1\%. 
This supports transfer beyond the training vocabulary. The SFT and GRPO data, objectives, and budgets differ, so the comparison does not establish a general ranking of training methods.

\paragraph{The competence profile still degrades.}
Figure~\ref{fig:grpo_curves} resolves the gains by chain length.
On front/behind, GRPO increases original/equivalent accuracy to 81.2/75.1\% and invariance to 73.3\%. 
At $n=256,512,1024$, its chance-corrected invariance is $-0.12,-0.11,-0.20$, respectively. 
Thus improved pooled performance and cross-vocabulary transfer coexist with substantial long-chain presentation failures. Training extends measured performance without meeting the competence condition across the tested range.

\section{Discussion and Diagnostic Evidence}
\label{sec:mechanism}
The competence profile distinguishes correct answers across presentations from agreement on an incorrect default. A higher aggregate score can reflect a wider successful range while substantial failures persist elsewhere in the family. The chain-of-thought and training studies illustrate this distinction.

Open-weight interventions provide supporting evidence. In Qwen3-8B, ablating selected heads raises invariance from 81.0\% to 93.0\% while original/equivalent accuracy falls from 64.5/59.5\% to 59.0/52.0\%. 
Stable answers can therefore accompany reduced task signal. In Qwen3.5-9B, short-chain answers remain linearly decodable despite biased outputs; longer-chain equivalents show reduced decodability at a measured layer and position. GRPO strengthens the measured signal relative to the answer default. 
These results characterize possible contributors without establishing a complete mechanism. Appendix~\ref{app:mechanism} reports the probes and interventions.

CBE turns such observations into a reusable testing strategy: state the competence, specify a family with controlled complexity, certify answer-preserving presentations, and report where correctness survives those changes. This makes model comparisons and intervention gains interpretable in terms of the behavior they actually support.

\section{Conclusion}
\label{sec:conclusion}
We proposed \emph{competence-based evaluation} as a methodology for evaluating LLM reasoning through a problem family. Its five components make the target competence and its necessary behavioral condition explicit. The ordering instantiation tests reliable application of the same rule across growing inputs and equivalent presentations; it does not isolate an inference-depth limit.

Even strong models fail that condition across the tested ordering family, despite reasonably good performance over bounded complexity ranges. Accuracy alone conceals additional deterioration and symmetric changes across equivalent presentations. Chain-of-thought extends successful performance, and GRPO improves held-out transfer under the tested recipes, but documented failures remain. Better performance therefore does not establish competence across the family.

These profiles can guide model development by locating gains across sizes and presentations, while benchmark designers extend coverage under the same competence condition.

The intended impact is an evaluation practice that grows with models: generate fresh instances, extend the complexity range, and test the same competence under matched presentations. CBE provides a reusable way to distinguish an expanded range of success from a demonstrated ability to handle the task throughout the evaluated family.

\clearpage
\section*{Limitations}
\label{sec:limitations}
CBE evaluates competence within a specified problem family. Finite samples can reveal failures but cannot certify correctness throughout an unbounded family. Our ordering study and Horn-clause adoption example leave broader reasoning competences untested. Extending CBE to richer tasks and wider presentation coverage would clarify how its diagnostic profiles generalize.

Chain length measures input size and presentation load; endpoint queries admit a shortcut. Our ordering results therefore do not isolate multistep inference capacity. Length and presentation changes are also not always separated. Future generators could independently control derivation depth, distracting information, and linguistic realization.

Behavioral profiles and linear probes do not uniquely identify internal reasoning processes. Paired metrics also condition on valid answers, so agreement must be read alongside completion coverage. Broader sampling and causal interventions could connect systematic behavioral failures to their mechanisms.

The reasoning and training interventions studied here leave many routes to improved competence unexplored. Their gains do not establish a general ranking of training methods, particularly when data and budgets differ. Matched comparisons across training objectives, model families, and compute budgets would help identify which approaches extend correct, invariant behavior beyond the trained range.

\section*{Ethical Considerations}
\label{sec:ethics}
CBE is intended to support research on reasoning competence. A potential societal risk is treating success on a generated family as evidence of general reasoning ability or readiness for deployment in consequential settings. Conversely, failures on this family should not be taken as evidence that a model lacks all useful reasoning abilities. We mitigate these interpretation risks by specifying the competence condition and evaluation coverage, reporting correctness alongside invariance, and keeping conclusions scoped to the tested family. Repeated optimization against a known generator may also encourage benchmark-specific strategies; fresh instances alone do not establish transfer to other reasoning tasks.

The reported ordering instances assign fictional relations to generic first-name labels or random strings, and gold answers are computed from the generated structure. These relations do not describe identifiable people's actual attributes. No human participants or crowdworker annotations are used to establish the task labels. The English templates and name vocabulary do not constitute demographic or multilingual coverage. Although some templates concern attributes such as wealth or seniority, they should not be interpreted as claims about real people or population groups.

Model inference, extended reasoning traces, and adapter training consume computational resources and energy. We reuse pretrained models and train LoRA adapters, but do not claim that these choices establish a low environmental footprint. We report evaluation and training settings in Appendix~\ref{app:details}; aggregate energy consumption and carbon emissions were not measured. Reporting competence gains alongside their computational cost is an important direction for future evaluations.

\bibliography{colm2026_conference}

\appendix

\section{Proofs and Statistical Notes}
\label{app:proofs}
\subsection{Paired-Outcome Identities}

For the common valid-pair support, write $A_k$ for accuracy on presentation $k$ and $\bar A=(A_0+A_1)/2$. The following elementary results clarify what paired observations add to marginal scores.

\begin{result}[Exact decomposition]
\label{res:decomposition}
$\bar A=R+U/2=\tfrac12+(R-D)/2$, and feasible $(\iota,\bar A)$ pairs satisfy $|\bar A-\tfrac12|\leq\iota/2$.
\end{result}

\begin{result}[Marginals do not identify paired competence]
\label{res:identifiability}
Given only $(A_0,A_1)$, robust correctness ranges over $[\max(0,A_0+A_1-1),\min(A_0,A_1)]$. Every value in this interval is attainable. A paired flip rate identifies $U=1-\iota$, but alone does not split agreement into $R$ and $D$. Labelled matched pairs identify all three masses.
\end{result}

\paragraph{Proof of Result~\ref{res:decomposition}.}
Answers are binary and both presentations share the gold label, so each matched pair realizes exactly one of the events $R$ (both correct), $D$ (both wrong) or $U$ (exactly one correct), and $R + D + U = 1$. The random variable $(C_0 + C_1)/2$ equals $1$ on $R$, $\tfrac12$ on $U$ and $0$ on $D$; taking expectations gives $\bar A = R + U/2$. Substituting $U = 1 - R - D$ gives $\bar A = \tfrac12 + (R - D)/2$. Since $\iota = R + D$ and $R, D \geq 0$, $|R - D| \leq \iota$, hence $|\bar A - \tfrac12| \leq \iota/2$; the three vertices of the triangle are the pure regimes (all robust-correct, all shared-wrong, all presentation-dependent). Beyond binary answers, with $U$ still meaning ``exactly one correct,'' $\bar A = R + U/2$ still holds and agreement is at most $R + D$, so $R \geq \bar A - (1 - \iota)/2$ remains a conservative bound. \qed

\paragraph{Proof of Result~\ref{res:identifiability}.}
$R = \Pr(C_0 = 1, C_1 = 1)$ is a functional of the joint law of $(C_0, C_1)$, whereas $A_0 = \Pr(C_0 = 1)$ and $A_1 = \Pr(C_1 = 1)$ are its marginals. For fixed marginals, the joint probability of $(1, 1)$ ranges over the Fr\'echet--Hoeffding interval $[\max(0, A_0 + A_1 - 1), \min(A_0, A_1)]$, and each endpoint and every value between them is attained by a $2 \times 2$ table with those marginals (vary the $(1,1)$ cell and adjust the other three). Any score that is a function of $(A_0, A_1)$ therefore constrains $R$ to that interval and no further. A paired flip rate observes only whether the two answers agree, i.e., $\iota = R + D$, and the two tables $(R, D) = (\iota, 0)$ and $(0, \iota)$ with the same $U$ are indistinguishable to it. Labelled pairs observe $(C_0, C_1)$ directly and so identify all three masses. \qed

\paragraph{Two different chance references.}
The tables use the pooled output-bias reference $\iota_e=q^2+(1-q)^2$, where $q$ is the pooled yes-rate over both presentations within a complexity level. This corresponds to independent answers with a common marginal rate $q$. If the two presentations have different yes-rates $q_0$ and $q_1$, the independence reference with their separate output marginals is instead $q_0q_1+(1-q_0)(1-q_1)$. The difference between the pooled and separate references is $(q_0-q_1)^2/2$. Thus the pooled convention additionally penalizes differing output biases. Neither convention supplies a correctness-based covariance interpretation.

For correctness indicators, define the separate reference $e_C=A_0A_1+(1-A_0)(1-A_1)$. Since $D=1-A_0-A_1+R$,
\[
\iota-e_C=2(R-A_0A_1)=2\,\mathrm{Cov}(C_0,C_1).
\]
This identity concerns $e_C$, not the output-bias reference used in the tables. Balanced gold labels do not make the two references equal: a perfect predictor on a balanced task has $e_C=1$, while $q=1/2$ gives $\iota_e=1/2$. We therefore do not read the sign of the reported $\kappa$ as the sign of correctness covariance. If $q$ is zero or one, the reported normalization is undefined. All profile identities apply on the same valid-pair support.

\paragraph{Why the boundary is not reported as a threshold crossing.}
Let $\hat n^*(\tau)$ denote the smallest $n$ at which $\hat\iota_n$ falls below a threshold $\tau$. A parametric bootstrap (binomial resampling of each $\hat\iota_n$ at the observed sample size, $B = 4{,}000$) on the Protocol L base pair shows that the probability of reproducing the observed crossing at $\tau = 0.85$ in a fresh dataset of 200 pairs per $n$ is $0.50$ for GPT-5 and $0.39$ for GPT-4o on front/behind, and remains $0.54$--$0.64$ for GPT-5 even at 400--3{,}200 pairs per $n$ because its curve sits close to $\tau$ at the crossing, whereas it is $0.87$--$1.00$ for models whose curves cross steeply (GPT-4.1, GPT-4.1-mini, Qwen3.5-9B). 
A curve that happens to sit on the threshold makes the crossing a coin flip at any sample size. The shape of the profile, by contrast, is stable under resampling, which is why the main text describes where profiles drop steeply or diverge and summarizes with $\bar\kappa$.

\paragraph{Sample sizes.}
At 200 pairs per $n$ the 95\% interval on $\iota_n$ has half-width $\pm 6.9$ points at $\iota = 0.5$ and $\pm 4.9$ at $\iota = 0.85$; the overall $\iota$ over the nine $n$ levels (1{,}800 pairs) has half-width $\pm 2.3$ points, and a paired difference in $\iota$ between two arms on the same instances of about $3.7$ points is detectable at 80\% power. 
Per-cell accuracy within one $n$ (50 instances) has half-width up to $\pm 14$ points, which is why per-cell polarity claims are pooled over $n$.

\section{Generators, Protocols, and Training Details}
\label{app:details}
For reproducibility, this appendix names the archived presentation conditions: Protocol P denotes the position-uniform datasets used in the base evaluation, while Protocol L denotes the release datasets, including the matched GPT-5.4 reasoning-effort comparison. These labels identify dataset construction, not different competence criteria. All effort levels in a condition use the same paired prompts.

\subsection{Ordering Generator}
\label{app:dataset}

\paragraph{Entity pool and premises.}
For the \textsc{pos} relation, names are sampled without replacement from a pool of about $2{,}500$ first names; a random-name control replaces them with random five-letter strings. For each instance the generator samples $n$ names $a_0, \ldots, a_{n-1}$, emits the $n - 1$ adjacent comparisons $a_i \prec a_{i+1}$ rendered in the relation's forward phrase, and shuffles both the premise order and the names list independently of the chain. The query is the endpoint pair $(a_0, a_{n-1})$ with direction and gold answer balanced over the four cells.

\paragraph{Prompt format.}
A representative Protocol P prompt at $n = 8$, verbatim:
\begin{quote}\small\ttfamily
There are 8 people standing in some order.\\
Their names are ['Heather', 'Charles', 'Joseph', 'James', 'Andrew', 'Michelle', 'Mark', 'Donna'].\\
Their standing order can be described by the following set of rules:\\
James is in front of Andrew.\\
Joseph is in front of James.\\
Mark is behind Donna.\\
Heather is in front of Charles.\\
Donna is in front of Joseph.\\
Andrew is in front of Michelle.\\
Charles is in front of Mark.\\
Is Heather in front of Michelle? Provide your answer only as yes or no. Answer:
\end{quote}
The system prompt, identical for every model and for training, is ``You are a helpful assistant. Answer logical reasoning questions concisely.''

\paragraph{Equivalent presentation under the two protocols.}
Under both protocols the equivalent presentation rewrites $m$ of the $n - 1$ premises with the inverse phrase and swapped arguments and re-shuffles the premise order and the names list. Under Protocol L (the release generator), $m$ is drawn uniformly from $\{\max(1, \lfloor 0.2(n-1) \rfloor), \ldots, n-1\}$ and the rewritten premises are the last $m$ comparisons in chain order for $n \leq 1024$, while at $n = 2048$ each premise is rewritten independently with probability $\nicefrac12$. Under Protocol P the same instances, names lines, premise orders and rewrite counts are kept and the rewritten positions are redrawn uniformly over the chain, so that relation phrasing is uncorrelated with chain position. On the same instances, equivalent-presentation accuracy is lower under Protocol P than under Protocol L in 43 of 50 (model, vocabulary) cells, by up to $7.7$ points (GPT-5 on before/after, $90.5 \to 82.8$); invariance is lower by $2$--$4$ points for GPT-5 on seven of eight arms and for GPT-4.1 on four of eight, and within noise for the mini models, both Qwen models and the LoRA adapter on all but two arms; original-presentation accuracy is shared by construction. 
The per-$n$ trends in Appendix~\ref{app:legacy} that cross $n = 1024 \to 2048$ therefore cross a change in the presentation distribution and are not interpreted. The release code ships both samplers.

\paragraph{Catalog.}
The 31 relations are rendered from the seed-1 instances with the same orderings, name-to-position assignment, names line, premise order, rewrite count and rewrite positions as \textsc{pos} (871 consistency checks; the \textsc{pos} files are byte-identical). Families and phrases: spatial line (in front of/behind, ahead of/behind, left/right, right/left, behind/in front of, before/after, north/south, east/west, closer to the front/back than, further forward/back than); vertical (above/below, higher up/lower down than); temporal (earlier/later to arrive, born before/after, earlier/later to finish, scheduled before/after); comparative (older, taller, shorter, heavier, faster, stronger, richer, each with its antonym); rank and social (more senior/junior, ranked higher/lower, more/less experienced, more/less popular, paid more/less, closer to/farther from the exit); nonce (more/less zorbic than; vorpward/morpward of, with the converse rule stated in the context). 

\subsection{Evaluation Protocol}
\label{app:protocol}

\paragraph{Decoding.}
Provider-default sampling was used throughout (OpenAI chat completions at temperature $1.0$; Qwen models served by vLLM at temperature $0.6$, top-$p$ $0.95$, top-$k$ $20$). Completion caps: $20$ tokens for GPT-4o, GPT-4o-mini, GPT-4.1, GPT-4.1-mini and the Qwen models (thinking disabled); the GPT-5 family at \texttt{reasoning\_effort=minimal} with \texttt{max\_completion\_tokens}$=1024$ (observed reasoning tokens zero); GPT-5.4 at \texttt{none} with the same cap and at \texttt{low}/\texttt{medium}/\texttt{high} with a $32{,}768$-token cap ($65{,}536$ for \texttt{medium} and \texttt{high} in the $n = 2048$ block); Qwen3.5-27B with thinking enabled at a $24$k budget. Identical-prompt repeat agreement and its interpretation are reported in Appendix~\ref{app:repeat_agreement}. Unparseable rates are below $1\%$ for every Protocol P model except Gemma-4-31B (Appendix~\ref{app:extra_models}).

The matched GPT-5.4 effort comparison uses Protocol L equivalents. With Protocol P equivalents, the pooled \texttt{none} row is 77.1\% original accuracy, 74.9\% equivalent accuracy, 78.1\% invariance, and $\bar\kappa=0.40$. 

\paragraph{Statistics.}
McNemar uses all-prompt correctness, counting invalid answers wrong, with the continuity-corrected $\chi^2$ for $b+c>25$ and the exact binomial test otherwise. Holm correction is across the compared models. Bootstrap intervals carry all presentations of a matched instance together; the reported catalog and GRPO contrasts use $2{,}000$ resamples stratified by relation. The new robust-correct curves use pointwise $95\%$ Wilson intervals on the same jointly parsed support as all three plotted quantities. Figure~\ref{fig:competence_profiles} is recomputed from archived answers, with no new model calls. Its source table records valid-pair counts and input-file hashes. 

\subsection{Training Details}
\label{app:training}

\paragraph{Supervised adapters.}
The cross-relation LoRA fine-tune of Qwen3.5-9B uses $45{,}600$ chat-formatted examples from $5{,}700$ orderings ($950$ per relation, $n \leq 128$) over six auxiliary relations (arrival order, priority, proximity, seniority, left/right, up/down), each ordering contributing its four query cells under both presentations; \textsc{pos} is held out. LoRA rank $16$, $\alpha = 32$, dropout $0.05$, all attention and MLP projections, AdamW at $2 \times 10^{-4}$ with cosine schedule and $0.03$ warm-up, one epoch, effective batch $16$, bf16, \texttt{max\_length} $1536$, seed $42$, one GPU. The presentation-pair SFT and the consistency-loss variant (a Jensen--Shannon term between the answer distributions of the presentations of an instance, with a chance-corrected guard) use the same pool. The identical recipe was applied to Qwen3.5-27B.

\paragraph{GRPO.}
GRPO adapters train on batches holding every presentation of an instance (original, rewritten, reshuffled, fully rewritten) with $G = 4$ samples per prompt, learning rate $10^{-5}$, $\beta = 0$, $600$ instances at $n \in [16, 128]$, $600$ steps, and seeds $42$ and $43$. Training uses $19$ relations without spatial-line vocabulary, of which \texttt{above/below}, \texttt{taller/shorter} and \texttt{higher/lower} share words with three evaluated relations that are therefore excluded from the clean held-out set. 

For a parsed answer $i$, the implemented reward is
\begin{equation}
\begin{aligned}
r_i&=c_i+\tfrac12 b_i,\\
b_i&=\max\{0,a_i-[q_z^2+(1-q_z)^2]\},
\end{aligned}
\end{equation}
where $c_i=\mathbf{1}[\operatorname{parse}(o_i)=y^*(z)]$, $a_i$ is the fraction of parsed answers on the instance's other presentations matching answer $i$, and $q_z$ is the yes-rate among all parsed samples of the instance. Unparseable answers receive zero reward. For each prompt $p$, advantages are
\begin{equation}
A_i=\frac{r_i-\mu_p}{\sigma_p+10^{-4}},
\end{equation}
where $\mu_p$ and $\sigma_p$ are the mean and standard deviation of its four sampled rewards. Appendix~\ref{app:grpo} analyzes the reward components under this normalization and reports the additional training seed. 

\section{Adopting CBE for Horn-Clause Inference}
\label{app:horn}
\subsection{Mapping the Framework to a New Task}
\label{sec:gen_horn}

This appendix illustrates adoption of CBE beyond the ordering benchmark. The latent generator is a Horn program over pronounceable pseudoword atoms. An exact forward-chaining oracle supplies the gold answer to a goal-derivability query. The complexity parameter is the number of rules, $n\in\{4,8,16,32,64,128\}$. 
Positive programs derive the goal; negative near misses retain a rule headed by the goal but sever its derivation at one point. Gold answers are balanced.
Rule count indexes program size. Positive proof depth and distractor content co-vary with rule count, so these profiles do not isolate proof-depth effects.

The two renderers express each rule either as “If $a$ and $b$, then $c$” or as “$c$ holds whenever $a$ and $b$ hold”. Rule order stays fixed. The model and competence criterion are unchanged: both answers must be correct. The same $(R,D,U)$ profile, invariance, and chance correction apply. Thus adoption requires task-specific latent instances, an oracle, and semantics-preserving renderers, without changing the outcome decomposition or analysis. This demonstrates a second instantiation; it does not measure implementation effort or imply that arbitrary presentation equivalence can be certified automatically.

\subsection{Illustrative Results}
\label{sec:horn}

\begin{table}[t]
\centering
\scriptsize
\setlength{\tabcolsep}{3pt}
\begin{tabular}{lrrrrrrr}
\toprule
Model & Acc.Org & Acc.Eqv & Inv. & $\bar\kappa$ & $R$ & $D$ & $U$ \\
\midrule
GPT-4o       & 60.2 & 61.7 & 75.0 & 0.46 & 48.5 & 26.6 & 25.0 \\
GPT-4o-mini  & 55.2 & 53.9 & 79.2 & 0.46 & 44.2 & 35.1 & 20.8 \\
GPT-4.1      & 62.5 & 60.7 & 79.8 & 0.41 & 51.5 & 28.3 & 20.2 \\
GPT-4.1-mini & 60.5 & 58.6 & 74.1 & 0.47 & 46.6 & 27.5 & 25.9 \\
GPT-5        & 60.0 & 60.8 & 73.0 & 0.40 & 46.9 & 26.1 & 27.0 \\
GPT-5-mini   & 57.0 & 58.3 & 72.2 & 0.37 & 43.8 & 28.4 & 27.8 \\
Qwen3.5-27B  & 65.2 & 63.6 & 82.0 & 0.51 & 55.4 & 26.6 & 18.0 \\
Qwen3.5-9B   & 52.4 & 52.8 & 68.4 & 0.09 & 36.8 & 31.6 & 31.6 \\
\bottomrule
\end{tabular}
\caption{Horn-clause inference pooled over $n\in\{4,\dots,128\}$ rules (Protocol L, 200 pairs per $n$, pairs whose two answers both parse). Accuracy (\%) on both presentations, per-pair invariance, $\bar\kappa$, and the joint readout $R$ / $D$ / $U$ (both correct / both wrong / exactly one correct, \%). No re-sampling floor was measured in this run.}
\label{tab:horn_summary}
\end{table}

Eight models receive 200 pairs per rule count under Protocol L; no identical-prompt retest was collected. 
The paired profile again reveals information beyond marginal accuracy. Pooled original/equivalent accuracy differs by less than two points for every model, while $U$ ranges from 18.0\% to 31.6\%. 
At $n=4$, six models have $\kappa_n$ between 0.67 and 0.94. By $n=32$, all eight models have original accuracy within four points of chance. 
Qwen3.5-27B has the largest pooled robust-correct mass, $R=55.4\%$, alongside $D=26.6\%$ and $U=18.0\%$. 
Its 82.0\% invariance therefore includes substantial shared error. 
Table~\ref{tab:horn_per_n} resolves the three masses by rule count. These results are an adoption example with narrower protocol coverage than the main ordering study.

\begin{table*}[t]
\centering
\small
\begin{tabular}{llrrrrrr}
\toprule
Model & & $n=4$ & $n=8$ & $n=16$ & $n=32$ & $n=64$ & $n=128$ \\
\midrule
GPT-4o & $R$ & 82.5 & 61.5 & 40.8 & 34.9 & 34.0 & 36.4 \\
 & $D$ & 7.0 & 21.0 & 27.0 & 28.2 & 36.0 & 40.4 \\
 & $U$ & 10.5 & 17.5 & 32.1 & 36.9 & 29.9 & 23.2 \\
 & $\kappa_n$ & 0.78 & 0.60 & 0.31 & 0.26 & 0.34 & 0.48 \\
\midrule
GPT-4o-mini & $R$ & 55.5 & 50.5 & 40.5 & 38.5 & 35.0 & 45.0 \\
 & $D$ & 26.5 & 35.0 & 41.0 & 33.5 & 37.0 & 37.5 \\
 & $U$ & 18.0 & 14.5 & 18.5 & 28.0 & 28.0 & 17.5 \\
 & $\kappa_n$ & 0.48 & 0.55 & 0.51 & 0.43 & 0.41 & 0.39 \\
\midrule
GPT-4.1 & $R$ & 90.0 & 55.0 & 39.5 & 36.0 & 44.5 & 44.0 \\
 & $D$ & 1.5 & 19.5 & 28.5 & 32.5 & 44.0 & 44.0 \\
 & $U$ & 8.5 & 25.5 & 32.0 & 31.5 & 11.5 & 12.0 \\
 & $\kappa_n$ & 0.83 & 0.49 & 0.33 & 0.19 & 0.35 & 0.30 \\
\midrule
GPT-4.1-mini & $R$ & 82.5 & 51.0 & 36.5 & 38.0 & 38.0 & 33.5 \\
 & $D$ & 8.0 & 20.0 & 32.0 & 32.5 & 35.5 & 37.0 \\
 & $U$ & 9.5 & 29.0 & 31.5 & 29.5 & 26.5 & 29.5 \\
 & $\kappa_n$ & 0.80 & 0.42 & 0.31 & 0.41 & 0.47 & 0.41 \\
\midrule
GPT-5 & $R$ & 79.5 & 58.5 & 46.5 & 37.0 & 33.0 & 27.0 \\
 & $D$ & 8.0 & 26.0 & 27.5 & 30.5 & 29.5 & 35.0 \\
 & $U$ & 12.5 & 15.5 & 26.0 & 32.5 & 37.5 & 38.0 \\
 & $\kappa_n$ & 0.74 & 0.54 & 0.36 & 0.29 & 0.24 & 0.24 \\
\midrule
GPT-5-mini & $R$ & 69.5 & 50.5 & 39.0 & 32.0 & 35.5 & 36.0 \\
 & $D$ & 16.0 & 30.5 & 28.0 & 32.5 & 33.0 & 30.5 \\
 & $U$ & 14.5 & 19.0 & 33.0 & 35.5 & 31.5 & 33.5 \\
 & $\kappa_n$ & 0.67 & 0.46 & 0.30 & 0.26 & 0.32 & 0.22 \\
\midrule
Qwen3.5-9B & $R$ & 55.0 & 42.5 & 35.2 & 37.0 & 24.7 & 26.1 \\
 & $D$ & 19.0 & 34.5 & 38.7 & 34.5 & 34.3 & 28.6 \\
 & $U$ & 26.0 & 23.0 & 26.1 & 28.5 & 40.9 & 45.2 \\
 & $\kappa_n$ & 0.31 & 0.04 & -0.02 & 0.12 & 0.06 & 0.03 \\
\midrule
Qwen3.5-27B & $R$ & 95.5 & 69.0 & 46.5 & 42.5 & 40.0 & 39.0 \\
 & $D$ & 1.5 & 13.5 & 28.5 & 35.0 & 42.0 & 39.0 \\
 & $U$ & 3.0 & 17.5 & 25.0 & 22.5 & 18.0 & 22.0 \\
 & $\kappa_n$ & 0.94 & 0.61 & 0.38 & 0.32 & 0.46 & 0.36 \\
\bottomrule
\end{tabular}
\caption{Horn-clause inference by number of rules $n$ (Protocol L, 200 pairs per $n$). $R$ / $D$ / $U$: percentage of jointly parseable pairs with both presentations correct / both wrong / exactly one correct. Chance correction uses $\kappa_n = (\iota_n - \iota_e)/(1 - \iota_e)$ with $\iota_n=R+D$ and $\iota_e=q^2+(1-q)^2$, where $q$ is the pooled yes-rate on the same support. GPT-4o and Qwen3.5-9B lose a few pairs to unparseable answers. No retest control was collected.}
\label{tab:horn_per_n}
\end{table*}

\section{Per-Axis Comparison with Prior Work}
\label{app:positioning}

Tables~\ref{tab:rw_compact_grouped} and~\ref{tab:rw_compact_additional} compare the designs behind Section~\ref{sec:positioning} along generation, complexity control, presentation changes, and diagnostic readouts. The descriptions distinguish tests of equivalent presentations of a fixed instance from changes to the answer, query, or reasoning problem. Several predecessors provide parameterized generation or paired consistency and correctness measures, including ParaRel Consistent-Acc. CBE builds on these contributions by connecting matched complexity profiles to an explicit behavioral condition on a problem family.

\begin{table*}[t]
\centering
\footnotesize
\setlength{\tabcolsep}{3pt}
\renewcommand{\arraystretch}{1.12}
\begin{tabular}{p{0.25\textwidth}p{0.34\textwidth}p{0.36\textwidth}}
\toprule
\textbf{Work} & \textbf{Generation and complexity} & \textbf{Presentations and diagnostic readout} \\
\midrule
CheckList \citep{ribeiro2020checklist} & Templates, lexicons and perturbation tools construct behavioral tests. & Capability-specific tests include invariance and directional expectations, with failure rates. \\
\addlinespace
Contrast sets \citep{gardner2020contrast} & Experts minimally edit existing examples to probe local decision boundaries. & Edits typically change the gold label; contrast-set performance tests sensitivity to meaningful changes. \\
\addlinespace
Metamorphic testing / LGMT \citep{chen1998metamorphic,segura2016survey,zhou2026lgmt} & Follow-up cases are derived through specified metamorphic relations. & LGMT uses logical equivalences and cross-case consistency checks, with analyses by transformation type. \\
\addlinespace
ParaRel / BECEL \citep{elazar2021pararel,jang2022becel} & Paraphrase patterns and task-specific transformations generate related inputs. & ParaRel includes correctness across all paraphrases; BECEL distinguishes consistency types and paired inconsistency. \\
\addlinespace
ReClor / LogiQA \citep{yu2020reclor,liu2020logiqa} & Curated examination questions form fixed released datasets. & Accuracy analyses include question types or ReClor's EASY/HARD subsets. \\
\addlinespace
FOLIO / P-FOLIO \citep{han-etal-2024-folio,han-etal-2024-p} & Human-annotated arguments; P-FOLIO adds annotated reasoning chains. & Entailment, translation to logic, proof generation and inference-rule analyses provide distinct evaluation views. \\
\addlinespace
LogicBench / Multi-LogiEval \citep{parmar2024logicbench,patel2024multilogieval} & LogicBench targets individual inference rules; Multi-LogiEval composes rules at several depths. & Rule- and depth-specific accuracy; controlled depth variation is explicit in Multi-LogiEval. \\
\addlinespace
PrOntoQA \citep{saparov2023greedy} & Synthetic first-order worlds support control over proof structure. & Generated reasoning is parsed into symbolic proof steps for analysis. \\
\addlinespace
ZebraLogic \citep{pmlr-v267-lin25i} & Generated logic-grid puzzles vary constraints and search-space complexity. & Accuracy and reasoning strategies are analyzed across puzzle complexity. \\
\addlinespace
DivLogicEval \citep{chung2025divlogiceval} & Solver-verifiable propositional structures are instantiated with diverse, counterintuitive sentences. & Accuracy, option-rotation checks and PartialCircular examine answer reliability. \\
\addlinespace
LogicSkills \citep{rabern2026logicskills} & Solver-verified arguments use a restricted fragment of first-order logic. & Symbolization, countermodels and validity are evaluated in English and a nonce-word language. \\
\addlinespace
ReCogLab \citep{liu2025recoglab} & Configurable relational tasks vary entity count, reasoning demands, premise order and filler text. & Scaling and cognitive effects include presentation order and symbolic distance. \\
\bottomrule
\end{tabular}
\caption{Comparison of generation, complexity control, presentation changes and diagnostic readouts in the cited evaluation designs. Descriptions identify the kind of change being tested, including whether the answer or query changes. Further comparisons appear in Table~\ref{tab:rw_compact_additional}.}
\label{tab:rw_compact_grouped}
\end{table*}

\begin{table*}[t]
\centering
\footnotesize
\setlength{\tabcolsep}{3pt}
\renewcommand{\arraystretch}{1.12}
\begin{tabular}{p{0.25\textwidth}p{0.34\textwidth}p{0.36\textwidth}}
\toprule
\textbf{Work} & \textbf{Generation and complexity} & \textbf{Presentations and diagnostic readout} \\
\midrule
RobustLR \citep{sanyal2022robustlr} & Deductive inputs undergo minimal logical edits and equivalence transformations. & Logical-operator and equivalence-specific robustness analyses. \\
\addlinespace
Premise Order Matters \citep{chen2024premise} & Deductive premises and GSM8K problems are reordered. & Answer-preserving reordering tests sensitivity to premise order. \\
\addlinespace
The Reversal Curse \citep{berglund2024reversal} & Synthetic-fact fine-tuning is evaluated using reverse-direction factual questions. & Tests generalization from a trained relation to reverse queries, which change the query direction. \\
\addlinespace
The Illusion of Thinking \citep{shojaee2025illusion} & Controllable puzzle environments vary problem complexity. & Final answers and reasoning traces are examined across complexity regimes. \\
\addlinespace
LiveBench \citep{xu2024livebench} & Regularly refreshed questions draw on recent sources and objective answers. & Task- and category-specific scores; freshness is distinct from a fixed-family complexity parameter. \\
\addlinespace
\textit{Mirage} \citep{schaeffer2023emergent} & Reanalyzes model scaling and evaluation metrics. & Studies how metric choice changes the appearance of abrupt capability gains. \\
\addlinespace
DyVal \citep{zhu2024dyval} & A programmatic directed-acyclic-graph generator controls reasoning-task complexity. & Scores and prompting strategies are evaluated across generated difficulty levels. \\
\addlinespace
GSM-Plus \citep{li2024gsmplus} & Existing mathematical problems are augmented with several perturbation types. & Robustness under changes that may preserve or alter the required answer. \\
\addlinespace
Paraphrase and Solve \citep{zhou-etal-2024-paraphrase} & Generates paraphrases of existing mathematical problems. & Studies surface-form sensitivity and combines samples across paraphrases. \\
\addlinespace
GSM-Symbolic \citep{mirzadeh2024gsmsymbolic} & Symbolic templates generate name and number variants, with additional clause manipulations. & Variation across instantiations and arithmetic load; number changes can change the answer. \\
\addlinespace
MATH-Perturb \citep{huang2025math} & Constructs simple and hard perturbations of existing mathematical problems. & Hard perturbations deliberately change the solution pattern, extending beyond answer-preserving reformulation. \\
\addlinespace
\textbf{CBE (this paper)} & Parameterized ordering generation fixes the task rule while varying input size. & Requires correctness across matched answer-preserving presentations and tracks accuracy and invariance by size. \\
\bottomrule
\end{tabular}
\caption{Further comparison of the evaluation designs described in the cited sources. Generating fresh inputs, varying problem complexity, testing presentation changes and reporting diagnostic measurements have multiple precedents. CBE connects these elements through its specified competence condition and matched profiles.}
\label{tab:rw_compact_additional}
\end{table*}

\clearpage
\section{Additional Models and Per-$n$ Tables}
\label{app:per_n}
\subsection{Additional Open-Weight Models on a Nine-Relation Core}
\label{app:extra_models}

Three further open-weight models were run under Protocol P on a nine-relation core (front/behind, left/right, north/south, above/below, born before/after, taller, shorter, ranked higher/lower, nonce direction): Gemma-4-31B, Qwen3.6-35B-A3B (3B active parameters) and Qwen2.5-32B-Instruct. Table~\ref{tab:core_models} gives family means on the matched core. Two observations. First, the local ladder is not monotone in model generation: Qwen2.5-32B exceeds Qwen3.5-27B on invariance and $\bar\kappa$ in every family and on original accuracy in every family but temporal, with the highest re-sampling floors (95--98) and the largest deficit relative to its retest agreement ($-12.9$ over the core, $-23$ on north/south, where original accuracy $90$ falls to $74$ on the equivalent). Second, Gemma-4-31B does not fail the task so much as abandon the answer format: under the 20-token cap it begins a chain of thought instead of answering on $29/84\%$ (original/equivalent) of front/behind pairs at $n = 128$ and $36/68\%$ at $n = 256$, so its invariance over parsed pairs ($96$--$98$) hides a strict invariance of $14$--$44$ at $n = 128$, and its accuracy columns measure format compliance; its parsed answers are $93$--$100\%$ correct at every $n \leq 128$ on every lexical relation, and its greedy first-token readout is $96$--$97\%$ correct at $n = 512$ with the gold answer decodable from the residual stream at $0.94$--$1.00$ through $n = 512$ in the probed layers. A free-form condition with a $1{,}024$-token budget changes almost nothing. All four fail the nonce direction words ($\bar\kappa \leq 0.03$). 

\begin{table}[t]
\centering
\scriptsize\setlength{\tabcolsep}{2.3pt}
\begin{tabular}{lrrrrrr}
\toprule
Model & Acc.\,Org & Acc.\,Eqv & Inv. & Inv.\,strict & Inv.$-$floor & $\bar\kappa$ \\
\midrule
Qwen3.5-27B      & 82.9 & 77.1 & 74.5 & 74.5 & $-6.7$  & 0.43 \\
Qwen3.5-9B       & 66.2 & 65.4 & 62.7 & 62.4 & $-4.7$  & 0.18 \\
Gemma-4-31B      & 86.1 & 77.5 & 91.6 & 72.9 & $-7.8$  & 0.82 \\
Qwen3.6-35B-A3B  & 67.0 & 66.5 & 63.3 & 63.1 & $-4.1$  & 0.18 \\
Qwen2.5-32B      & 86.5 & 83.6 & 83.5 & 83.5 & $-12.9$ & 0.59 \\
9B + GRPO     & 89.4 & 87.6 & 87.1 & 87.1 & $-8.8$  & 0.68 \\
\bottomrule
\end{tabular}
\caption{Means over the matched nine-relation core (Protocol P). Inv.\,strict counts a pair with an unparseable answer as a disagreement; Gemma's gap between the two columns is format abandonment under the 20-token cap. The last row is the correctness-only GRPO adapter of \S\ref{sec:training}.}
\label{tab:core_models}
\end{table}

\subsection{Presentation Components, Distractors, and Negation}

Table~\ref{tab:forms} separates the components of the equivalent presentation for the two flagship GPT models under Protocol P, and Table~\ref{tab:dose_negation} reports the distractor-dose and negation runs under Protocol L (\S\ref{sec:forms}).

\begin{table}[t]
\centering
\footnotesize
\setlength{\tabcolsep}{4pt}
\begin{tabular}{lcc}
\toprule
Equivalent presentation & GPT-4.1 & GPT-5 \\
\midrule
identical prompt re-sent (floor) & 93.4 / 96.4 & 95.0 / 93.7 \\
\midrule
premise order reshuffled only     & 93.2 / 92.6 & 93.6 / 91.1 \\
25\% of premises re-worded        & 92.6 / 91.3 & 89.3 / 87.8 \\
50\% re-worded                    & 90.3 / 89.9 & 84.8 / 85.5 \\
75\% re-worded                    & 90.5 / 90.1 & 86.1 / 86.1 \\
100\% re-worded                   & 84.1 / 80.2 & 82.4 / 81.4 \\
100\% re-worded $+$ reshuffled    & 84.1 / 79.7 & 82.9 / 81.2 \\
\bottomrule
\end{tabular}
\caption{What the presentation change is made of: Acc.Eqv / Inv.\ (\%) on the front/behind instances of Protocol P when the equivalent presentation changes only the premise order, only the fraction of premises re-worded with the converse relation (order fixed), or both. Original-presentation accuracy is 93.1 (GPT-4.1) and 94.1 (GPT-5) in every row. The first row re-sends the original prompt unchanged.}
\label{tab:forms}
\end{table}

\begin{table}[t]
\centering
\footnotesize
\setlength{\tabcolsep}{4pt}
\begin{tabular}{lrrrrr}
\toprule
 & \multicolumn{3}{c}{distractor dose $\rho$} & \multicolumn{2}{c}{presentation} \\
\cmidrule(lr){2-4}\cmidrule(lr){5-6}
Model & 0 & 0.5 & 1 & negation & retest \\
\midrule
GPT-4.1      & 93.2 & 91.1 & 89.6 & --   & --   \\
GPT-4.1-mini & 76.0 & 76.2 & 74.3 & --   & --   \\
GPT-5        & 92.9 & 91.0 & 89.8 & 90.0 & 97.0 \\
GPT-5-mini   & 68.5 & 66.0 & 65.8 & 64.1 & 78.9 \\
Qwen3.5-27B  & 75.2 & 72.7 & 69.6 & 60.2 & 81.8 \\
Qwen3.5-9B   & 59.0 & 55.5 & 55.0 & 56.0 & 69.5 \\
\bottomrule
\end{tabular}
\caption{Per-pair invariance (\%) pooled over $n\in\{4,\dots,256\}$ under Protocol L (see Section~\ref{sec:instantiation}). Left: the syntactic equivalent with $\mathrm{round}(\rho\,(n-1))$ distractor premises over disjoint entities added to both presentations. Right: the same items and premise subsets with each re-worded premise expressed as the negation of its converse, and the agreement of the original prompt with itself re-sent (retest). The $\rho=0$ column is the syntactic family.}
\label{tab:dose_negation}
\end{table}

\subsection{Output Decomposition at the Answer Position}

\begin{figure}[t]
\centering
\includegraphics[width=0.85\linewidth]{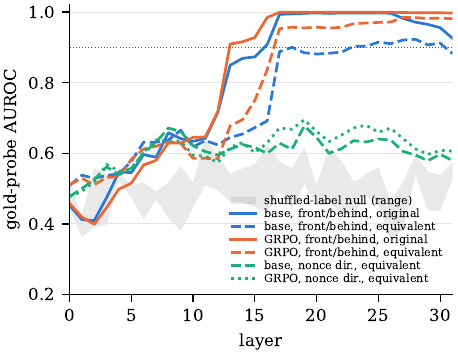}
\caption{Gold-probe AUROC by layer at the answer position, Qwen3.5-9B (Protocol P instances, greedy readout, $n \le 32$, 200 pairs per relation): base and correctness-only GRPO on front/behind, original (solid) and equivalent (dashed), and the equivalent presentation of the nonce direction words. Grey band: shuffled-label null; dotted line: AUROC 0.9.}
\label{fig:probes}
\end{figure}

Table~\ref{tab:probes} gives the output decomposition behind \S\ref{sec:mechanism}, and Figure~\ref{fig:probes} the per-layer gold-probe curves.

\begin{table}[t]
\centering
\footnotesize\setlength{\tabcolsep}{2pt}
\resizebox{\linewidth}{!}{%
\begin{tabular}{lrrrrr}
\toprule
Relation & Acc. & Yes & $\delta$ & $s$ & Layer \\
\midrule
\multicolumn{6}{l}{\emph{Qwen3.5-9B, base}} \\
front/behind & 55.5/59.5 & 96/92 & 1.58/1.14 & 0.63/0.50 & 16/18 \\
left/right & 69.5/68.5 & 82/84 & 1.67/1.41 & 1.65/1.38 & 13/13 \\
nonce dir. & 56.0/55.5 & 96/88 & 1.19/0.72 & 0.57/0.14 & 17/-- \\
\midrule
\multicolumn{6}{l}{\emph{Qwen3.5-9B, correctness-only GRPO}} \\
front/behind & 87.5/87.0 & 64/58 & 1.37/0.36 & 3.60/3.22 & 13/17 \\
left/right & 97.5/99.0 & 55/53 & $-$0.22/$-$0.46 & 7.64/7.09 & 13/13 \\
nonce dir. & 90.5/60.0 & 56/36 & $-$0.06/$-$1.39 & 3.59/1.00 & 16/-- \\
\midrule
\multicolumn{6}{l}{\emph{Qwen3-8B, base}} \\
front/behind & 64.5/59.5 & 82/86 & 2.42/3.52 & 3.54/2.33 & 19/22 \\
\bottomrule
\end{tabular}}
\caption{Output decomposition at the answer position (Protocol P instances, $n \le 32$, 200 pairs per relation, greedy readout). Each cell is original/equivalent presentation. Acc.\ and yes-rate (Yes) in \%; $\delta$ = presentation-level default toward ``yes'' and $s$ = instance signal, in logits; Layer = first layer at which the cross-validated gold probe reaches AUROC 0.9 (-- = never; the Qwen3-8B replicate has 36 layers, the 9B 32). Nonce dir.\ = nonce direction words.}
\label{tab:probes}
\end{table}

\subsection{Reasoning Effort by Chain Length}

\begin{figure*}[t]
\centering
\includegraphics[width=0.8\textwidth]{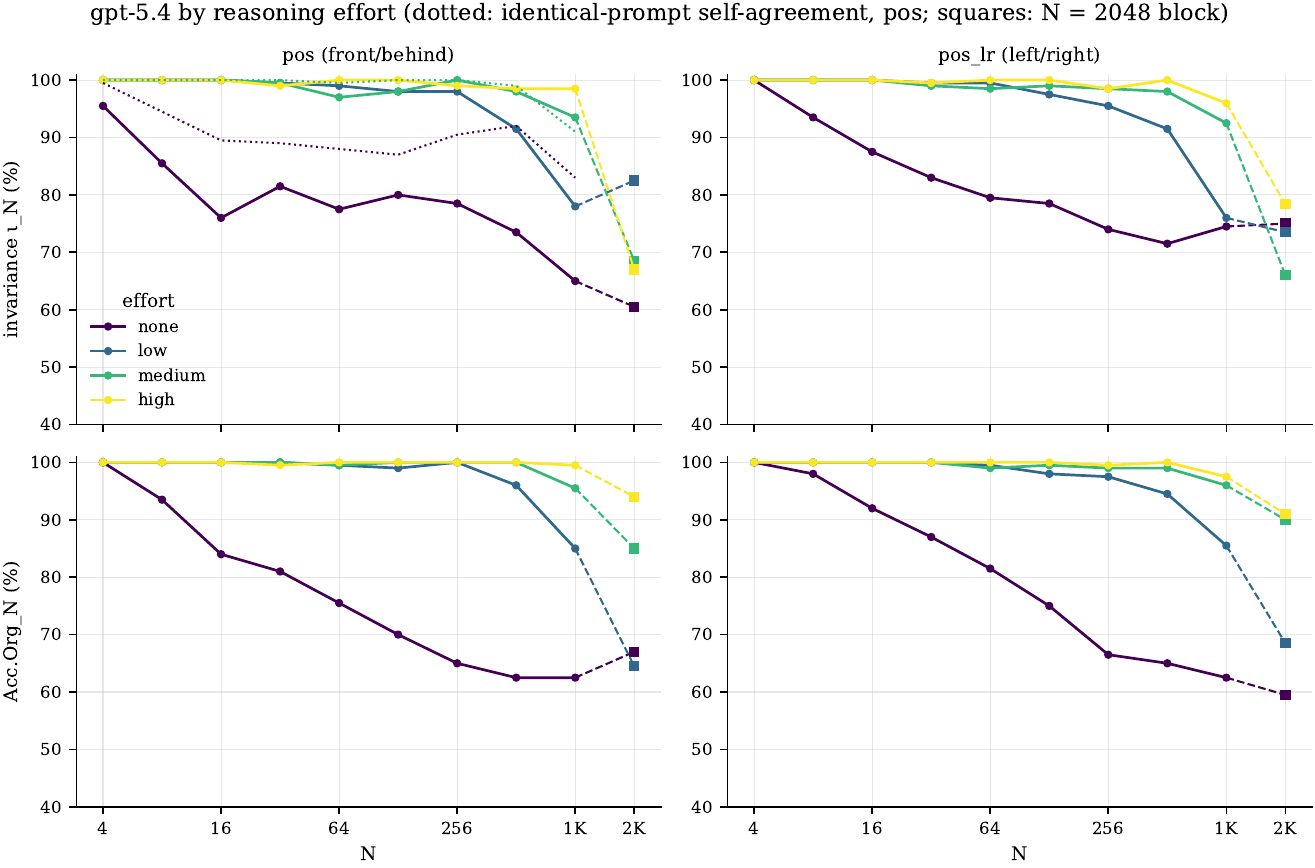}
\caption{GPT-5.4 by reasoning effort: per-pair invariance (top) and original-presentation accuracy (bottom) by $n$ under front/behind (left) and left/right (right). Dotted: identical-prompt self-agreement; squares: the $n=2048$ block. The \texttt{none} curves use the Protocol P instances with the Protocol L equivalents; \texttt{low}, \texttt{medium}, \texttt{high} and the $n=2048$ block are Protocol L equivalents on the same instances.}
\label{fig:reasoning}
\end{figure*}

\subsection{Per-$n$ Tables}

\begin{figure*}[t]
\centering
\includegraphics[width=0.9\textwidth]{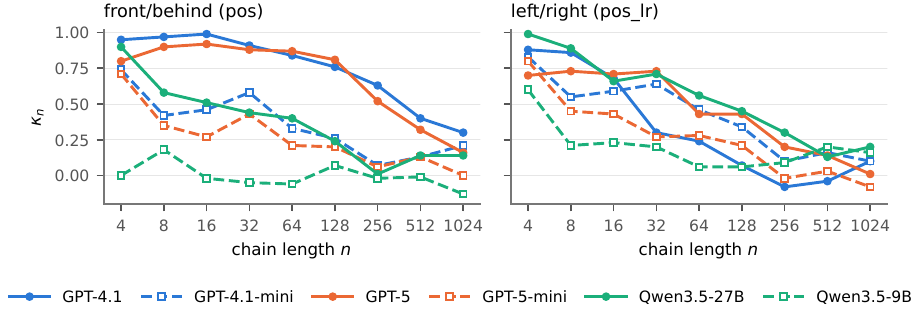}
\caption{Chance-corrected invariance $\kappa_n$ by chain length for the six base models under front/behind (left) and left/right (right), Protocol P. Solid lines denote the larger and dashed lines the smaller model of each family. The dotted reference is $\kappa_n=0$ under the pooled output-bias convention.}
\label{fig:kappa_per_n}
\end{figure*}

Figure~\ref{fig:kappa_per_n} gives chance-corrected invariance for both base vocabularies. Table~\ref{tab:per_n_pos} gives per-$n$ accuracy on both presentations and invariance for the six Protocol P models on front/behind. Horn-clause profiles are in Appendix~\ref{app:horn}.

\begin{table*}[t]
\centering
\footnotesize\setlength{\tabcolsep}{4pt}
\begin{tabular}{llrrrrrrrrr}
\toprule
Model & & 4 & 8 & 16 & 32 & 64 & 128 & 256 & 512 & 1024 \\
\midrule
\multirow{3}{*}{GPT-4.1-mini}& Acc.\,Org & 95.5 & 87.5 & 91.0 & 84.0 & 75.5 & 69.0 & 61.5 & 59.5 & 59.5 \\
& Acc.\,Eqv & 90.5 & 76.0 & 78.0 & 81.5 & 71.0 & 71.5 & 61.0 & 51.0 & 62.0 \\
& Inv. & 87.0 & 72.5 & 74.0 & 79.5 & 72.5 & 69.5 & 65.5 & 64.5 & 67.5 \\ \midrule
\multirow{3}{*}{GPT-4.1}& Acc.\,Org & 98.5 & 100.0 & 99.5 & 99.5 & 97.0 & 94.0 & 89.5 & 82.0 & 77.5 \\
& Acc.\,Eqv & 98.0 & 98.5 & 100.0 & 96.0 & 95.0 & 92.5 & 86.0 & 74.0 & 68.5 \\
& Inv. & 97.5 & 98.5 & 99.5 & 95.5 & 92.0 & 87.9 & 81.5 & 71.0 & 67.3 \\ \midrule
\multirow{3}{*}{GPT-5-mini}& Acc.\,Org & 94.0 & 75.5 & 68.0 & 70.0 & 69.0 & 67.0 & 61.5 & 63.5 & 61.5 \\
& Acc.\,Eqv & 91.5 & 79.5 & 79.0 & 74.5 & 71.0 & 70.0 & 67.5 & 61.5 & 53.5 \\
& Inv. & 85.5 & 69.0 & 70.0 & 77.5 & 66.0 & 61.0 & 53.0 & 59.0 & 63.0 \\ \midrule
\multirow{3}{*}{GPT-5}& Acc.\,Org & 96.5 & 99.5 & 99.0 & 99.5 & 98.0 & 97.5 & 92.0 & 89.0 & 76.0 \\
& Acc.\,Eqv & 92.5 & 95.5 & 97.0 & 93.5 & 95.5 & 92.0 & 78.0 & 71.5 & 68.0 \\
& Inv. & 90.0 & 95.0 & 96.0 & 93.9 & 93.5 & 90.5 & 76.8 & 68.4 & 67.0 \\ \midrule
\multirow{3}{*}{Qwen3.5-9B}& Acc.\,Org & 67.0 & 58.5 & 56.0 & 57.0 & 54.5 & 55.0 & 54.5 & 47.5 & 50.5 \\
& Acc.\,Eqv & 62.0 & 55.0 & 56.5 & 55.0 & 54.0 & 52.0 & 45.0 & 54.0 & 46.0 \\
& Inv. & 57.0 & 75.3 & 71.9 & 62.3 & 51.5 & 55.0 & 49.2 & 49.5 & 44.5 \\ \midrule
\multirow{3}{*}{Qwen3.5-27B}& Acc.\,Org & 100.0 & 93.5 & 94.0 & 89.5 & 84.5 & 78.0 & 71.0 & 63.0 & 59.0 \\
& Acc.\,Eqv & 95.0 & 83.5 & 79.5 & 78.0 & 73.5 & 69.0 & 61.0 & 61.0 & 56.0 \\
& Inv. & 95.0 & 79.0 & 76.5 & 73.5 & 72.0 & 63.0 & 53.0 & 64.0 & 73.0 \\
\bottomrule
\end{tabular}
\caption{Front/behind under Protocol P, per $n$ (200 pairs each): accuracy on the original and equivalent presentations and per-pair invariance (\%).}
\label{tab:per_n_pos}
\end{table*}

\section{Results under the Release Protocol}
\label{app:legacy}
This appendix collects the results obtained under Protocol L (the benchmark release: eight models, $400$ pairs per $n$, $n \leq 2048$, no re-sampling floor; Appendix~\ref{app:dataset}). They are the basis of an earlier version of this work and are reported here, with the figures recomputed from the released outputs, because the eight-model suite includes the GPT-4o family and the $n = 2048$ level, neither of which was re-run under Protocol P. Comparisons across the two protocols are not drawn.

\subsection{Eight Models on the Release Protocol}

\begin{figure*}[t]
    \centering
    \includegraphics[width=0.8\textwidth]{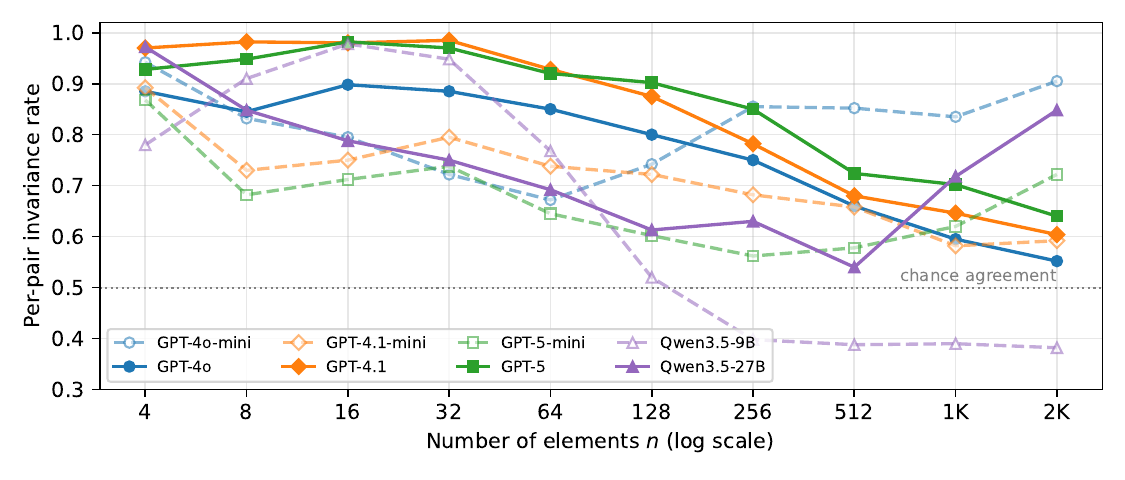}
    \caption{Per-pair invariance by $n$ for the eight models under Protocol L (front/behind, 400 pairs per $n$). Dotted line: chance agreement.}
    \label{fig:legacy_invariance}
\end{figure*}

Table~\ref{tab:all_models_pos_summary} summarizes accuracy and invariance pooled over $n \in \{4, \ldots, 2048\}$, Table~\ref{tab:all_models_pos_mcnemar} the McNemar tests, and Table~\ref{tab:all_models_pos_invariance_per_n} the per-$n$ invariance; Figure~\ref{fig:legacy_invariance} plots the latter. The qualitative picture agrees with \S\ref{sec:base_pair}: GPT-4.1 and GPT-5 keep invariance above $87\%$ through $n = 128$ and fall steeply afterwards, every other model is at or below $85\%$ by $n = 8$ except Qwen3.5-9B, and high invariance at large $n$ for GPT-4o-mini and Qwen3.5-27B coincides with both presentations defaulting to ``no'' (accuracy $45$--$55\%$), i.e., shared-wrong mass rather than competence. Qwen3.5-9B's $91$--$98\%$ agreement at $n = 8$--$32$ coincides with $50\%$ accuracy and is a yes-default; at $n \geq 256$ its agreement falls below chance. Six of eight models reject the marginal null at Holm-corrected $p < 0.001$, yet the per-$n$ discordant counts change sign within rows for GPT-4o and GPT-4.1-mini (original favoured at small $n$, equivalent at $n = 128$ or $512$--$1024$), which the aggregate cancels. Seven of eight models prefer the backward query at $n = 2048$ by $6$--$23$ points, and six of eight collapse onto a ``no'' default there, with terminal polarity gaps from about $10$ points (GPT-4o) to $83$ (Qwen3.5-27B).

\begin{table}[t]
\centering
\small\setlength{\tabcolsep}{4pt}
\begin{tabular}{lrrrr}
\toprule
Model & Acc.\,Org & Acc.\,Eqv & \(\Delta\) & Inv. \\
\midrule
GPT-4o-mini & 71.7 & 67.5 & +4.2 & 81.5 \\
GPT-4o & 83.2 & 82.3 & +0.8 & 77.2 \\
GPT-4.1-mini & 71.6 & 74.1 & -2.5 & 71.4 \\
GPT-4.1 & 87.4 & 90.8 & -3.4 & 84.4 \\
GPT-5-mini & 66.6 & 74.1 & -7.5 & 67.3 \\
GPT-5 & 90.6 & 88.3 & +2.2 & 85.7 \\
Qwen3.5-9B & 56.9 & 63.3 & -6.5 & 64.6 \\
Qwen3.5-27B & 77.3 & 72.1 & +5.2 & 74.0 \\
\bottomrule
\end{tabular}
\caption{Per-model accuracy and invariance on the \textsc{pos} dataset (\(N\!\in\!\{4,8,\ldots,2048\}\), 4000 questions per split). \textbf{Acc.\,Org} and \textbf{Acc.\,Eqv} are accuracy on original and equivalent splits; \(\Delta\) is their difference; \textbf{Inv.} is the fraction of paired questions where the model gave the same yes/no on both phrasings (over pairs with parseable answers). All values are percentages.}
\label{tab:all_models_pos_summary}
\end{table}

\begin{table*}[t]
\centering
\begin{tabular}{lrrrrrrr}
\toprule
Model & \(b\) & \(c\) & Gap & 95\% CI & \(\chi^2\) & \(p_{\text{McNemar}}\) & \(p_{\text{Holm}}\) \\
\midrule
GPT-4o-mini & 453 & 285 & +4.2 & [+2.9, +5.5] & 37.8 & \textless\,0.001 & \textless\,0.001 \\
GPT-4o & 473 & 439 & +0.9 & [-0.7, +2.3] & 1.2 & 0.27 & 0.27 \\
GPT-4.1-mini & 522 & 621 & -2.5 & [-4.1, -0.9] & 8.4 & 0.0037 & 0.0075 \\
GPT-4.1 & 247 & 384 & -3.4 & [-4.7, -2.2] & 29.3 & \textless\,0.001 & \textless\,0.001 \\
GPT-5-mini & 505 & 803 & -7.4 & [-9.3, -5.7] & 67.4 & \textless\,0.001 & \textless\,0.001 \\
GPT-5 & 332 & 243 & +2.2 & [+1.0, +3.4] & 13.5 & \textless\,0.001 & \textless\,0.001 \\
Qwen3.5-9B & 579 & 837 & -6.5 & [-8.3, -4.6] & 46.6 & \textless\,0.001 & \textless\,0.001 \\
Qwen3.5-27B & 624 & 417 & +5.2 & [+3.6, +6.7] & 40.8 & \textless\,0.001 & \textless\,0.001 \\
\bottomrule
\end{tabular}
\caption{McNemar's test on the original/equivalent accuracy gap, per model on the \textsc{pos} dataset (\(N=4000\) paired questions). \(b\) is the number of pairs with org correct and eqv wrong; \(c\) is the reverse. Gap = (org\_acc \(-\) eqv\_acc), with 95\% paired-bootstrap CI (\(B=10000\)). \(p_{\text{McNemar}}\) uses the continuity-corrected \(\chi^2\) statistic; \(p_{\text{Holm}}\) is the Holm-Bonferroni adjusted \(p\) across the 8 models.}
\label{tab:all_models_pos_mcnemar}
\end{table*}

\begin{table*}[t]
\centering
\begin{tabular}{lrrrrrrrrrr}
\toprule
Model & \multicolumn{10}{c}{\(N\)} \\
\cmidrule(lr){2-11}
 & 4 & 8 & 16 & 32 & 64 & 128 & 256 & 512 & 1024 & 2048 \\
\midrule
GPT-4o-mini & 94.2 & 83.2 & 79.5 & 72.2 & 67.2 & 74.2 & 85.5 & 85.2 & 83.5 & 90.5 \\
GPT-4o & 88.5 & 84.5 & 89.8 & 88.5 & 85.0 & 80.0 & 75.0 & 66.0 & 59.5 & 55.2 \\
GPT-4.1-mini & 89.2 & 73.0 & 75.0 & 79.5 & 73.8 & 72.2 & 68.2 & 65.8 & 58.2 & 59.2 \\
GPT-4.1 & 97.0 & 98.2 & 98.0 & 98.5 & 92.8 & 87.5 & 78.2 & 68.0 & 64.6 & 60.4 \\
GPT-5-mini & 86.8 & 68.2 & 71.2 & 73.8 & 64.5 & 60.2 & 56.2 & 57.8 & 62.0 & 72.2 \\
GPT-5 & 92.8 & 94.8 & 98.2 & 97.0 & 92.0 & 90.2 & 85.0 & 72.4 & 70.2 & 64.0 \\
Qwen3.5-9B & 78.0 & 91.0 & 97.8 & 94.8 & 76.8 & 52.0 & 39.8 & 38.8 & 39.0 & 38.2 \\
Qwen3.5-27B & 97.2 & 84.8 & 78.8 & 75.0 & 69.2 & 61.3 & 63.0 & 54.0 & 71.8 & 84.8 \\
\bottomrule
\end{tabular}
\caption{Per-model invariance on the \textsc{pos} dataset, by problem size \(N\). Invariance is the fraction of (original, equivalent) pairs receiving the same yes/no answer (over pairs with parseable answers). Values are percentages.}
\label{tab:all_models_pos_invariance_per_n}
\end{table*}

\subsection{Supervised Fine-Tuning under the Release Protocol}
\label{app:legacy_lora}

The cross-relation LoRA fine-tune of Qwen3.5-9B (Appendix~\ref{app:training}) evaluated on the held-out \textsc{pos} relation under Protocol L reduces the marginal gap from $6.5$ to $0.6$ points without raising accuracy ($60.1 \to 57.6$ mean), while its per-pair invariance moves from $64.6$ to $57.2$ and sits at chance ($48$--$50\%$) for every $n \geq 128$; the identical recipe on Qwen3.5-27B changes neither accuracy ($74.7$) nor the gap ($5.2 \to 5.6$). The marginal gap and the instance-level invariance thus move in opposite directions on the same checkpoint, which is Result~\ref{res:identifiability} observed in practice: an evaluation reporting only the marginal gap would read $|\Delta| = 0.6$ as near-perfect invariance. Under Protocol P the same adapter is within noise of its base on every axis (\S\ref{sec:training}).

\begin{table}[t]
\centering
\scriptsize\setlength{\tabcolsep}{3pt}
\begin{tabular}{lccccc}
\toprule
\textbf{Model} & \textbf{Acc.\,Org} & \textbf{Acc.\,Eqv} & \textbf{Mean} & $|\Delta|$ & \textbf{Inv.} \\
\midrule
Qwen3.5-9B (base)   & 56.9 & 63.3 & 60.1 & 6.5 & 64.6 \\
Qwen3.5-9B + LoRA   & 57.2 & 57.9 & 57.6 & 0.6 & 57.2 \\
Qwen3.5-27B (base)  & 77.3 & 72.1 & 74.7 & 5.2 & 74.0 \\
Qwen3.5-27B + LoRA  & 77.5 & 71.9 & 74.7 & 5.6 & 73.2 \\
\bottomrule
\end{tabular}
\caption{Cross-relation LoRA fine-tunes on the held-out \textsc{pos} relation under Protocol L, pooled over $n \in \{4, \ldots, 2048\}$. Values recomputed from the released outputs.}
\label{tab:legacy_lora}
\end{table}

\subsection{A Names-List Artifact as a Diagnostic}
\label{app:lora_leak}

An earlier generator emitted the names list in chain order, so that the list alone encoded the total order. A LoRA fine-tune of Qwen3.5-9B trained on such data reached $\geq 99\%$ accuracy at every $n$ up to $2048$ on both presentations when evaluated with the same generator, and $62.3 / 66.1$ (original/equivalent) once the names list was shuffled, about five points above the untouched base; the base Qwen3.5-27B itself scores $90.5$ on the original presentation with chain-ordered names against $77.3$ with shuffled names, so pretrained models pick up the cue zero-shot. The matched-presentation evaluation surfaced the discrepancy before the artifact was located, which is the methodological point: a presentation set does not need to know which surface feature a model exploits, only to disturb surface form while preserving logical content. All main ordering results use shuffled names; the chain-ordered condition here is a diagnostic control.

\begin{table}[t]
\centering
\scriptsize\setlength{\tabcolsep}{3pt}
\begin{tabular}{llccc}
\toprule
\textbf{Model} & \textbf{Names list} & \textbf{Acc.\,Org} & \textbf{Acc.\,Eqv} & \textbf{Inv.} \\
\midrule
Qwen3.5-9B (base)   & shuffled     & 56.9 & 63.3 & 64.6 \\
Qwen3.5-9B + LoRA   & chain order  & 100.0 & 99.9 & 99.9 \\
Qwen3.5-9B + LoRA   & shuffled     & 62.3 & 66.1 & 78.1 \\
Qwen3.5-27B         & shuffled     & 77.3 & 72.1 & 74.0 \\
\bottomrule
\end{tabular}
\caption{A fine-tune trained on chain-ordered names lists, evaluated with chain-ordered and with shuffled names lists (Protocol L, \textsc{pos}, pooled over $n$). The $78.1$ invariance under shuffled names coincides with a yes-default at large $n$ (accuracy on no-instances about $18\%$ at $n = 2048$).}
\label{tab:lora_leak_appendix}
\end{table}

\subsection{Cross-Relation Replication (\textsc{depth})}
\label{app:depth_table}

The same protocol applied to a \textsc{depth} relation (objects rather than people, $n \in \{4, \ldots, 64\}$, $2{,}000$ pairs per model) reproduces the model ordering of Table~\ref{tab:all_models_pos_summary} at comparable $n$, and the LoRA fine-tune again fails to raise accuracy over its base on this second held-out relation.

\begin{table}[t]
\centering
\footnotesize\setlength{\tabcolsep}{4pt}
\begin{tabular}{lrrrr}
\toprule
Model & Acc.\,Org & Acc.\,Eqv & \(\Delta\) & Inv. \\
\midrule
GPT-4o-mini & 94.6 & 94.7 & $-0.1$ & 91.3 \\
GPT-4o & 94.6 & 93.4 & $+1.2$ & 89.5 \\
GPT-4.1-mini & 87.9 & 92.6 & $-4.7$ & 83.9 \\
GPT-4.1 & 98.8 & 98.9 & $-0.1$ & 97.6 \\
GPT-5-mini & 86.5 & 87.0 & $-0.5$ & 78.2 \\
GPT-5 & 96.4 & 93.8 & $+2.7$ & 90.6 \\
Qwen3.5-9B & 82.3 & 89.0 & $-6.6$ & 83.0 \\
Qwen3.5-27B & 86.7 & 87.5 & $-0.8$ & 80.0 \\
Qwen3.5-9B + LoRA & 75.7 & 76.0 & $-0.3$ & 69.5 \\
\bottomrule
\end{tabular}
\caption{\textsc{depth} replication under Protocol L ($n \in \{4, \ldots, 64\}$).}
\label{tab:depth_summary}
\end{table}

\clearpage
\section{Additional GRPO Results and Reward Analysis}
\label{app:grpo}

\subsection{Training-Seed Replication}
\label{app:grpo_seeds}

\begin{figure*}[t]
\centering
\resizebox{\textwidth}{!}{\begingroup
\definecolor{curve0}{HTML}{666666}
\definecolor{curve1}{HTML}{0072B2}
\definecolor{curve2}{HTML}{D55E00}
\definecolor{curve3}{HTML}{009E73}
\begin{tikzpicture}
\begin{groupplot}[group style={group size=2 by 1,horizontal sep=36pt},
width=212pt,height=124pt,xmode=log,log basis x=2,
tick label style={font=\fontsize{8}{9}\selectfont},
label style={font=\fontsize{9}{10}\selectfont},
title style={font=\fontsize{9}{10}\selectfont},
xlabel={Chain length $n$},xmin=3.8,
xmax=1150,xtick={4,16,64,256,1024},xticklabels={4,16,64,256,1024},
axis lines=left,ymajorgrids=true,grid style={black!12},
legend style={font=\fontsize{8}{9}\selectfont,draw=none,at={(1.16,1.28)},
anchor=south,legend columns=4,column sep=6pt},
every axis plot/.append style={line width=.9pt,mark size=1.4pt}]
\nextgroupplot[ymin=40,ymax=103,ytick={50,75,100},title={Mean accuracy},ylabel={Accuracy (\%)}]
\addplot[color=curve0,dotted,mark=o] coordinates {(4,64.50000000) (8,57.32323232) (16,56.28140704) (32,56.03015075) (64,54.25000000) (128,53.50000000) (256,50.00000000) (512,50.75000000) (1024,48.25000000)};
\addlegendentry{Base 9B}
\addplot[color=curve1,dashed,mark=square*] coordinates {(4,66.83417085) (8,54.77386935) (16,53.51758794) (32,57.78894472) (64,55.61224490) (128,50.76142132) (256,55.77889447) (512,52.76381910) (1024,52.01005025)};
\addlegendentry{SFT + JS}
\addplot[color=curve2,dashdotted,mark=diamond*] coordinates {(4,92.50000000) (8,95.25000000) (16,91.00000000) (32,88.50000000) (64,84.00000000) (128,76.50000000) (256,63.50000000) (512,58.75000000) (1024,53.25000000)};
\addlegendentry{GRPO seed 42}
\addplot[color=curve3,solid,mark=triangle*] coordinates {(4,92.00000000) (8,87.25000000) (16,83.75000000) (32,77.00000000) (64,83.25000000) (128,79.25000000) (256,71.00000000) (512,67.00000000) (1024,56.25000000)};
\addlegendentry{GRPO seed 43}
\nextgroupplot[ymin=-.3,ymax=1.03,ytick={-.2,0,.5,1},title={Chance-corrected invariance},ylabel={$\kappa_n$}]
\draw[black!40,thin] (axis cs:4,0) -- (axis cs:1150,0);
\addplot[color=curve0,dotted,mark=o] coordinates {(4,0.00359170) (8,0.17706434) (16,-0.01715955) (32,-0.05235325) (64,-0.05575358) (128,0.06629318) (256,-0.01517792) (512,-0.01002525) (1024,-0.13383896)};
\addplot[color=curve1,dashed,mark=square*] coordinates {(4,0.16130268) (8,0.05068575) (16,0.00066076) (32,-0.05626327) (64,0.11127451) (128,-0.00685080) (256,-0.02577320) (512,-0.10555556) (1024,-0.08868267)};
\addplot[color=curve2,dashdotted,mark=diamond*] coordinates {(4,0.73871973) (8,0.82845178) (16,0.67793881) (32,0.59803035) (64,0.41626409) (128,0.18933333) (256,-0.12390351) (512,-0.11275696) (1024,-0.19794872)};
\addplot[color=curve3,solid,mark=triangle*] coordinates {(4,0.73480212) (8,0.64704939) (16,0.50384520) (32,0.38657833) (64,0.44827586) (128,0.32946486) (256,0.03846154) (512,-0.00816615) (1024,-0.24120185)};
\end{groupplot}
\end{tikzpicture}
\endgroup}
\caption{Training-seed replication on held-out front/behind (Protocol P, 200 pairs per chain length). Both GRPO seeds are compared with the base 9B and SFT with a JS consistency loss. Mean accuracy and $\kappa_n$ share valid-pair support. The main text displays seed 42, selected by mean invariance across the twelve clean held-out relations.}
\label{fig:grpo_seed_curves}
\end{figure*}
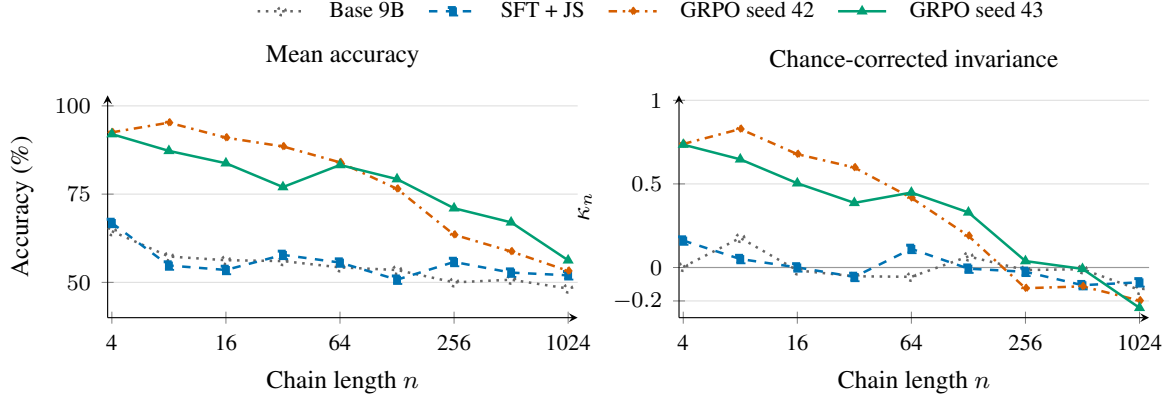

The main text reports seed 42, selected for higher mean invariance on the twelve clean held-out relations. This selection uses the reported held-out evaluation results. Seed 43 uses the same training and evaluation recipe. Table~\ref{tab:grpo_seeds} retains both runs, including their front/behind results, and Figure~\ref{fig:grpo_seed_curves} resolves their profiles by chain length. 
The selected run is not uniformly stronger on every metric: seed 43 has higher mean accuracy and $\bar\kappa$ across the twelve relations. Both runs retain substantial long-chain failures.

\begin{table}[t]
\centering
\footnotesize\setlength{\tabcolsep}{4pt}
\begin{tabular}{lrrrr}
\toprule
Run & Acc.Org & Acc.Eqv & Inv. & $\bar\kappa$ \\
\midrule
\multicolumn{5}{l}{\emph{Held-out front/behind}} \\
Seed 42 & 81.2 & 75.1 & 73.3 & 0.33 \\
Seed 43 & 78.1 & 76.7 & 67.3 & 0.32 \\
\midrule
\multicolumn{5}{l}{\emph{Means over twelve clean held-out relations}} \\
Seed 42 & 87.6 & 83.7 & 83.5 & 0.60 \\
Seed 43 & 87.9 & 85.6 & 81.3 & 0.61 \\
\bottomrule
\end{tabular}
\caption{GRPO training-seed replication (Protocol P, $n=4$ to $1024$, 200 pairs per relation and level). Both runs use the same recipe. Main-text results select seed 42 by mean invariance across the twelve clean held-out relations. Accuracy uses all prompts; paired metrics use jointly parsed answers. Percentages except $\bar\kappa$.}
\label{tab:grpo_seeds}
\end{table}

\paragraph{Why the agreement bonus was inert.}
The nominal reward of the GRPO runs in Section~\ref{sec:training} is $r = \mathrm{correct} + \lambda \max(0, \mathrm{agree} - \mathrm{chance})$ with $\lambda = 1/2$, where agree is the share of samples on the instance's other presentations that give the same answer and $\mathrm{chance} = q^2 + (1-q)^2$ with $q$ the instance's yes-rate. Each batch holds the four presentations of one instance (original, equivalent, re-shuffled, fully rewritten) times $G = 4$ samples, but the trainer forms the advantage group per prompt, that is per presentation. Within one group the other presentations' answers, $q$ and chance are shared by all four samples, and every answer is yes or no (no run produced an unparsed answer), so the reward takes two values, one for the gold answer and one for the other. The agreement term shifts the gap between them by at most $\lambda/2 = 0.25$ and cannot reverse it, since the correctness gap is 1, and with group-normalized advantages $(r - \mathrm{mean})/\mathrm{std}$ of a two-valued group depend only on how many of the four samples were right, not on the size of the gap. Hence $\lambda = 1/2$ and $\lambda = 0$ give the same advantages (up to the $10^{-4}$ added to the standard deviation), and the logged cosine between the advantage vector used and the $\lambda = 0$ vector is 1.0000 on every step with a nonzero advantage. The signal was also sparse: a step has a nonzero advantage only if some presentation's four samples disagree, which happened on 28 of 600 steps for GRPO seed 42 and 31 of 600 for seed 43, and the gradient is zero on every other step. The transfer in Section~\ref{sec:training} is the effect of these few correctness updates, not of an agreement term.

\paragraph{A variant in which the bonus can act.}
Rescaling the reward is not enough: the policy gradient is linear in the advantages and every nonzero GRPO step had a gradient norm of 2--38 against a clipping maximum of 1, so a bonus matters only if it changes the direction of the advantage vector. That requires all presentations of an instance to share one advantage group and the instance's rewards to take three or more values. The grpo-ig variant groups all sixteen samples of an instance, uses the instance's mean reward as the baseline without group scaling, and rewards $\mathrm{correct} + \lambda(\mathrm{agree} - \mathrm{chance})$ without the clip at zero; everything else follows the GRPO recipe (the same 19 held-in relations, 600 instances, $n = 16$ to $128$, $G = 4$). We trained $\lambda = 0$ and $\lambda = 0.5$ with seeds 42 and 43. Each $\lambda = 0.5$ run had 32--34 of 600 steps with a nonzero advantage, and the bonus changed the update direction on 30 of them, with a mean cosine of 0.9987 to the $\lambda = 0$ vector, about $3^\circ$: with yes/no answers and a gold label shared by all presentations, the majority answer of an informative instance is almost always the gold one, so rewarding agreement points the same way as rewarding correctness.

\begin{table}[t]
\centering
\footnotesize\setlength{\tabcolsep}{3pt}
\begin{tabular}{lrrrrr}
\toprule
 & \multicolumn{2}{c}{Acc.} & & & Inv.\ \\
Model & Org & Eqv & Inv. & $\bar\kappa$ & $-$floor \\
\midrule
Qwen3.5-9B, base & 61.4 & 60.2 & 60.0 & 0.10 & $-$5.3 \\
GRPO, seed 42 & 87.6 & 83.7 & 83.5 & 0.60 & $-$12.4 \\
GRPO, seed 43 & 87.9 & 85.6 & 81.3 & 0.61 & $-$14.2 \\
grpo-ig, $\lambda = 0$, seed 42 & 87.7 & 84.0 & 82.8 & 0.61 & $-$12.1 \\
grpo-ig, $\lambda = 0.5$, seed 42 & 88.8 & 84.6 & 84.6 & 0.63 & $-$10.7 \\
grpo-ig, $\lambda = 0$, seed 43 & 87.7 & 85.2 & 81.5 & 0.60 & $-$14.0 \\
grpo-ig, $\lambda = 0.5$, seed 43 & 88.2 & 85.6 & 81.7 & 0.62 & $-$13.0 \\
\bottomrule
\end{tabular}
\caption{Agreement bonus versus correctness-only GRPO on the twelve clean held-out relations (Protocol P, $n = 4$ to $1024$, 1,800 pairs per relation). Acc., Inv.\ and $\bar\kappa$ are means over the twelve relations; Inv.$-$floor over the eight relations with an identical-prompt floor. GRPO = per-presentation advantage groups (correctness-only in effect); grpo-ig = all sixteen samples of an instance in one advantage group, reward correct $+ \lambda(\text{agree} - \text{chance})$.}
\label{tab:grpo_ig}
\end{table}

Table~\ref{tab:grpo_ig} gives the twelve-relation evaluation. The $\lambda$ effect ($\lambda = 0.5$ minus $\lambda = 0$) is $+1.8$ points of invariance for seed 42 (95\% pair-bootstrap interval $+1.4$ to $+2.3$) and $+0.2$ for seed 43 ($-0.2$ to $+0.6$), with $\bar\kappa$ up by 0.02 for both seeds; the intervals cover evaluation noise only. Switching the training seed within one recipe moves invariance by 1.3--2.8 points (GRPO 2.2, grpo-ig $\lambda = 0$ 1.3, grpo-ig $\lambda = 0.5$ 2.8), so the $\lambda$ effect is within the seed-to-seed spread, and instance grouping on its own ($\lambda = 0$ against GRPO) changes invariance by $-0.7$ and $+0.2$. Every variant leaves the front/behind family as the weakest cells (invariance 65--79, $\bar\kappa$ 0.26--0.55) and the nonce direction words near zero chance-corrected invariance ($\bar\kappa$ $-0.08$ to $-0.01$). One seed shows an evaluation-level gain, but the two-seed evidence does not establish a consistent improvement beyond the observed training-seed variation.

\clearpage
\section{Identical-Prompt Repeat Agreement}
\label{app:repeat_agreement}
\subsection{Control and Interpretation}

We estimate \emph{identical-prompt repeat agreement} by sending each original prompt to the same model again under the same decoding settings and comparing the two answers. This measures run-to-run agreement without changing the prompt. We contrast it with invariance between the original and an equivalent presentation. Both agreement measures condition on parseable answers in the corresponding pair.

Earlier tables call this control the \emph{re-sampling floor} or self-agreement. It is a measured reference, not a universal lower bound. Cross-presentation agreement below it indicates sensitivity beyond the observed original-prompt repeat level. The difference does not causally isolate presentation error, since answer uncertainty can differ between presentations. It also does not establish a competence ranking by itself. Repeated sampling of both presentations would provide a stronger characterization of decoding variability.

\subsection{Base Ordering Models}

\begin{table}[t]
\centering
\small\setlength{\tabcolsep}{4pt}
\begin{tabular}{lrrr}
\toprule
Model & Inv. & Repeat & Inv.$-$Repeat \\
\midrule
GPT-4.1 & 87.9 & 96.4 & $-8.5$ \\
GPT-4.1-mini & 72.5 & 83.6 & $-11.1$ \\
GPT-5 & 85.7 & 93.7 & $-7.9$ \\
GPT-5-mini & 67.1 & 75.1 & $-8.0$ \\
Qwen3.5-27B & 72.1 & 78.0 & $-5.9$ \\
Qwen3.5-9B & 57.3 & 64.8 & $-7.5$ \\
\bottomrule
\end{tabular}
\caption{Base-model repeat controls on front/behind (Protocol P, $n=4$ to $1024$, 200 pairs per level). Inv. compares original and equivalent prompts; Repeat compares two responses to the identical original prompt. Values are percentages; differences are computed before rounding. Agreement conditions on parsed answers in each compared pair.}
\label{tab:repeat_base}
\end{table}

On front/behind, identical-prompt agreement ranges from 64.8\% to 96.4\% across the six base models. Cross-presentation invariance is 5.9--11.1 percentage points lower than the corresponding repeat agreement (Table~\ref{tab:repeat_base}). 
For GPT-4.1, the comparison is 96.4\% identical-prompt agreement against 87.9\% invariance. 
The output profiles in the main text therefore accompany a measured decoding control, rather than assuming perfectly repeatable responses.

Reshuffling-only agreement is 2.3--7.9 points below the repeat reference for the GPT models. 
This complements the small marginal accuracy changes reported in \S\ref{sec:forms}. Protocol L negation controls also show repeat agreement above syntactic and negated-presentation agreement for all four tested models; Qwen3.5-27B gives 81.8, 75.2, and 60.2\%, respectively. 
These observations describe sensitivity relative to the measured reference; they do not remove uncertainty induced by the altered prompts.

\subsection{Training Comparisons}

\begin{table}[t]
\centering
\small\setlength{\tabcolsep}{3pt}
\begin{tabular}{lrrr}
\toprule
Arm & Inv. & Repeat & Inv.$-$Repeat \\
\midrule
9B base & 57.3 & 64.8 & $-7.5$ \\
SFT, cross-relation & 56.6 & -- & -- \\
SFT, pairs & 58.0 & 65.1 & $-7.1$ \\
SFT, pairs + JS & 58.3 & 64.2 & $-5.8$ \\
GRPO, s42 & 73.3 & 93.7 & $-20.4$ \\
GRPO, s43 & 67.3 & 89.2 & $-21.9$ \\
\bottomrule
\end{tabular}
\caption{Repeat controls for the Qwen3.5-9B training arms on held-out front/behind (Protocol P, $n=4$ to $1024$, 200 pairs per level). Values are percentages; differences precede rounding. Dashes indicate an unavailable repeat control. Both SFT and GRPO use LoRA.}
\label{tab:repeat_training}
\end{table}

GRPO improves both repeat agreement and cross-presentation invariance, but by different amounts (Table~\ref{tab:repeat_training}). For seed-42 GRPO, front/behind repeat agreement rises from the base model's 64.8\% to 93.7\%, while invariance rises from 57.3\% to 73.3\%. The difference widens from $-7.5$ to $-20.4$ points; seed 43 gives $-21.9$. 
More repeatable original-prompt responses therefore coexist with substantial presentation failures. This comparison supports reporting both quantities, without attributing the residual disagreement wholly to a deterministic presentation-specific mechanism.

\subsection{Reasoning and Additional Models}

The available GPT-5.4 repeat controls are plotted in Figure~\ref{fig:reasoning}. On matched Protocol L front/behind prompts, \texttt{none} gives 90.3\% repeat agreement and 79.2\% invariance; \texttt{medium} gives 98.8\% and 98.4\%. 
Repeat controls were not collected for every reasoning-effort condition. Additional-model repeat comparisons remain in Appendix~\ref{app:extra_models}; the GRPO agreement-bonus comparisons are in Appendix~\ref{app:grpo}. Horn-clause runs have no repeat control.

\section{Mechanistic Probes and Interventions}
\label{app:mechanism}
Output profiles establish the behavioral failures. Probes and interventions in open-weight models provide supporting evidence about how presentation dependence arises, without identifying a complete mechanism.

\paragraph{Method.}
We study Qwen3.5-9B, its GRPO adapter (\S\ref{sec:training}), and Qwen3-8B with greedy yes/no readout. Short-chain runs use 200 matched pairs per relation at $n\le32$; complexity runs use 30 pairs per level through $n=256$. 
We fit cross-validated gold-answer probes \citep{belinkov2022probing}, apply the logit lens \citep{nostalgebraist2020lens}, and patch original residual states or attention heads into equivalent runs \citep{meng2022rome,vig2020causal}. A balanced-cell decomposition distinguishes the presentation-level default $\delta$ toward “yes” from instance signal $s$ (Table~\ref{tab:probes}).

\paragraph{Decodable answers and output defaults.}
At short chains, the base 9B's gold answer is linearly decodable on lexical relations (best probe AUROC 1.00 on originals, 0.92--1.00 on equivalents), despite nearly constant “yes” outputs. Front/behind has $\delta=1.58$ and $s=0.63$. 
Late residual patches often restore the original answer on disagreeing pairs (Figure~\ref{fig:probes}, Appendix~\ref{app:per_n}). These observations are consistent with a readout default overwhelming usable answer information. They do not rule out earlier representational errors.

\paragraph{Decodability weakens with complexity.}
At layer 26, equivalent-presentation AUROC drops from 0.98 at $n=16$ to 0.75 at $n=64$ and 0.54 at $n=256$ for the base model. At $n=256$, the adapter reaches 0.65 on equivalents, compared with 0.90 on originals. 
This demonstrates reduced linear decodability at the measured position and layer. It does not prove that the answer is absent from the model, since other representations or nonlinear readouts may retain it. Nonce direction terms also yield weak equivalent-presentation probes (base 0.68 against a shuffled-label null of 0.615). 
That evidence supports difficulty with this rendering, rather than a general claim that unfamiliar relations cannot be learned.

\paragraph{Agreement can rise when signal is removed.}
In Qwen3-8B, ablating four heads selected through cross-vocabulary patching removes about 40\% of $s$, compared with at most 3\% for the random-head controls. Invariance rises from 81.0\% to 93.0\% while original/equivalent accuracy falls from 64.5/59.5\% to 59.0/52.0\%. 
The increase in agreement accompanies worse task performance, directly illustrating why invariance alone is insufficient.

\paragraph{Training strengthens the measured signal.}
For front/behind, GRPO increases $s$ from 0.63 to 3.60 on originals and from 0.50 to 3.22 on equivalents. The default decreases from 1.58/1.14 to 1.37/0.36, remaining nonzero on originals. 
This is consistent with greater task signal relative to the default. These probes cover one model family, small datasets, and mostly short chains; they do not explain the API models' failures.

\end{document}